\documentclass{article} % For LaTeX2e
\usepackage{conference,times}

\usepackage{booktabs}
\usepackage{multicol}
\usepackage{multirow}
\usepackage{amsmath,amsfonts,bm}
\usepackage{xspace}

\usepackage[table,dvipsnames]{xcolor}

\usepackage{microtype}
\usepackage{cuted}
\usepackage{enumitem}
\usepackage{pifont}

\usepackage[most]{tcolorbox}
\usepackage{titletoc}
\usepackage{url}
\usepackage{subcaption}
\usepackage{tcolorbox}

\usepackage[
    colorlinks=true,
    citecolor=BlueViolet,
    linkcolor=purple,
    urlcolor=BlueViolet
]{hyperref}

\newcommand\DoToC{%
  \startcontents
  \printcontents{}{1}{\noindent \textbf{\Large{Table of Contents in Appendix}}\vskip3pt\vskip5pt}
  \vskip3pt\vskip5pt
}

\usepackage{amsmath,amsfonts,bm}

\def\eqref#1{equation~\ref{#1}}
\def\1{\bm{1}}

\DeclareMathAlphabet{\mathsfit}{\encodingdefault}{\sfdefault}{m}{sl}
\SetMathAlphabet{\mathsfit}{bold}{\encodingdefault}{\sfdefault}{bx}{n}

\usepackage{amssymb} % http://ctan.org/pkg/amssymbhttps://www.overleaf.com/project/6ab52785ec5c7e61116218c4#
\usepackage{bbm}   % for \mathbbm{1}; do NOT load bbold, it replaces the AMS \mathbb font
\usepackage{pifont} % http://ctan.org/pkg/pifont
\newcommand{\cmark}{\ding{51}}%
\newcommand{\xmark}{\ding{55}}%
\usepackage{wrapfig}
\usepackage{subcaption}
\usepackage{graphicx} %
\usepackage{sidecap, caption}
\usepackage{tikz} 
\usepackage{xcolor} 
\usepackage{multirow}
\usepackage{tocloft}
\usepackage{xspace}
\usepackage{enumitem}
\usepackage{bbding}

\definecolor{darkgreen}{rgb}{0,0.4,0}
\definecolor{codegreen}{rgb}{0,0.6,0}
\definecolor{carmine}{rgb}{0.59, 0.0, 0.09}

\newcommand{\ourmethod}{{{\sc FlowRig}}\xspace}
\newcommand\mypara[1]{\vspace{1mm}\noindent\textbf{#1}}

\usepackage{hyperref}       % hyperlinks
\usepackage{url}            % simple URL typesetting
\usepackage[utf8]{inputenc} % allow utf-8 input
\usepackage[T1]{fontenc}    % use 8-bit T1 fonts
\usepackage{booktabs}       % professional-quality tables
\usepackage{amsmath}        % advanced math typesetting
\usepackage{amsfonts}       % blackboard math symbols
\usepackage{nicefrac}       % compact symbols for 1/2, etc.
\usepackage{microtype}      % microtypography
\usepackage{xcolor}         % colors
\usepackage{graphicx}       % resize tables
\usepackage{multirow}       % multi-row cells in tables
\usepackage{enumitem}       % control list spacing and margins

\title{FlowRig: Unified Skeleton Generation for \\ 3D and 4D Meshes via Flow Matching}

\author{Fengzhi Guo \quad Nate Morrill \quad Shiqi Li \quad Cheng Zhang \\ [0.5em]
Texas A\&M University\\[0.5em]
\href{https://tamu-visual-ai.github.io/FlowRig/}{https://tamu-visual-ai.github.io/FlowRig/}
}

\iclrfinalcopy % Uncomment for camera-ready version, but NOT for submission.
\begin{document}

\maketitle

\begin{abstract}
Automatic rigging is a fundamental task for both static 3D and dynamic 4D mesh animation, yet existing methods do not provide a unified feedforward solution for these two settings. While static 3D riggers achieve strong quality, they often provide limited control over bone count and become temporally unstable when applied frame by frame to dynamic meshes. Existing 4D methods, in contrast, typically rely on expensive per-sequence optimization. We propose \ourmethod, a template-free, data-efficient framework for unified 3D and 4D skeleton generation. Built on flow matching, \ourmethod generates temporally aligned skeletons for an entire sequence in a single feedforward pass, while naturally handling single-mesh 3D rigging. The model further supports explicit skeleton-count control, uncertainty estimation for candidate selection, and masking-based interpolation, forecasting, and backcasting. To address limited paired 4D rigging data, we adopt a curriculum that initializes from large-scale 3D rigging data and adapts to mid-scale 4D mesh-skeleton data. Experiments show that \ourmethod achieves strong static 3D rigging performance with controllable bone complexity, as well as high-quality and temporally consistent 4D reconstruction and prediction results.
\end{abstract}

\begin{figure}[!h]
\centering
\includegraphics[width=\linewidth]{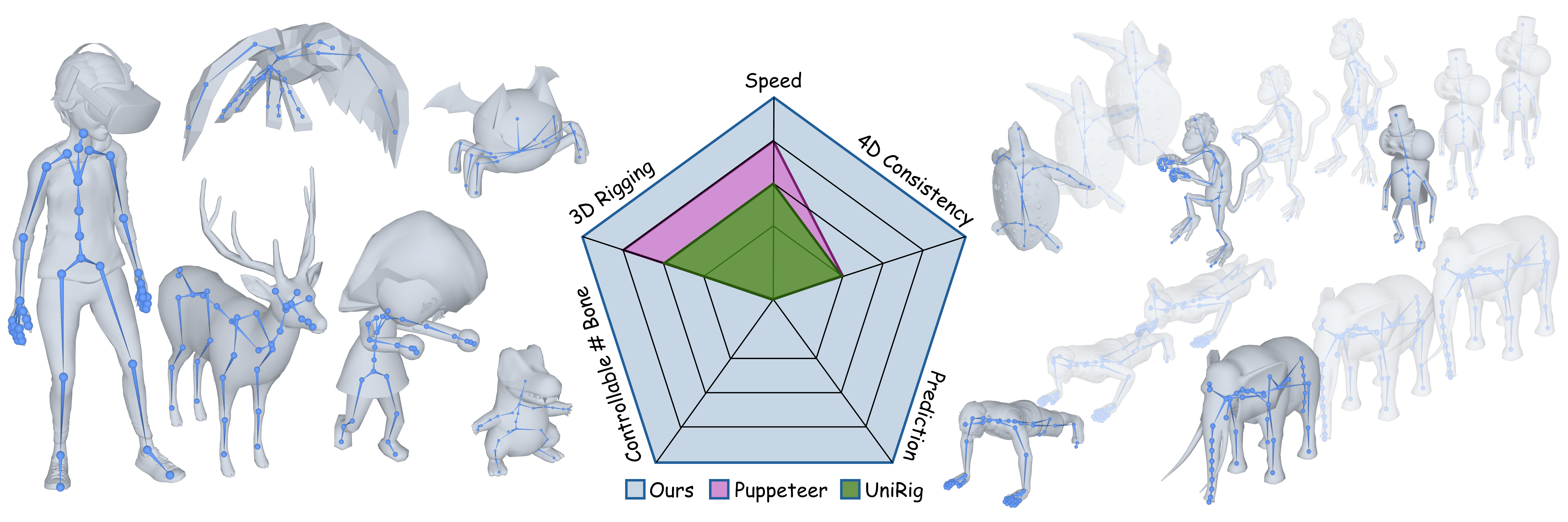}
\caption{\small \ourmethod unifies rigging for both 3D and 4D meshes. Compared with prior methods such as Puppeteer~\citep{song2025puppeteer} and UniRig~\citep{zhang2025one}, our model delivers accurate 3D rigging, preserves 4D temporal consistency, supports controllable bone count, achieves faster inference, and enables 4D skeleton prediction, including interpolation, forecasting, and backcasting.}
\label{fig:teaser}
\end{figure}

\section{Introduction}
Automatic rigging is a core problem in computer graphics, which turns raw geometry into animatable assets. This need spans both static 3D and dynamic 4D meshes: 3D assets require high-quality, controllable skeletons for downstream animation workflows, while 4D assets additionally require temporal consistency across time. At the same time, temporally coherent 4D meshes are becoming increasingly available from modern reconstruction and generation pipelines. Recent systems can recover or synthesize dynamic 4D geometry directly from videos or generative models, making 4D meshes a practical representation rather than a niche output~\citep{sabathier2026actionmesh, zhang2025texverse, xie2026animimic}. However, geometry alone is not enough for downstream animation, editing, simulation, and control. To make these assets truly useful, we need a unified feedforward rigging model that works for both 3D and 4D meshes, produces informative skeletons, supports controllable bone complexity, and remains stable across diverse poses and motions.

Unfortunately, this capability is still missing. On one side, existing rigging methods are predominantly designed for static 3D inputs, ranging from classical optimization systems~\citep{baran2007automatic} to modern feedforward and autoregressive models~\citep{xu2020rignet, zhang2025one, liu2025riganything, sun2025armo, song2025puppeteer, song2025magicarticulate, zhang2026rigmo}. These methods have substantially improved 3D rigging quality and efficiency, but they are fundamentally static, and many of them provide limited control over skeleton density or bone count for downstream applications. When applied frame by frame to dynamic meshes, they also typically fail to maintain temporal consistency~\citep{wang2026sprig}. On the other hand, existing 4D methods mainly focus on per-sequence optimization, reconstruction, or motion transfer~\citep{li2025articulated, yao2025riggs, wang2026gaussianimate, dabhi2026rat4d, muralikrishnan2025smf, kim2026morphgs}. While these methods are valuable, they do not provide a unified feedforward rigging solution that bridges controllable 3D rigging, dynamic 4D reconstruction, and predictive temporal reasoning within one framework.

Building such a unified model is challenging for two reasons. First, many strong static rigging methods are autoregressive, which makes explicit bone-count control less natural and global temporal reasoning inefficient for dynamic sequences. Second, paired 4D mesh-skeleton data remains far scarcer than static 3D rigging data, so a practical model must learn temporal consistency from limited 4D supervision while preserving strong 3D rigging quality.

In this work, we propose \ourmethod, a data-efficient feedforward framework for unified skeleton generation on 3D and 4D meshes. Built on flow matching, \ourmethod predicts temporally aligned skeletons for an entire sequence in one pass, while naturally handling 3D rigging with explicit skeleton-count control. The model also includes a graph head for temporally consistent topology prediction and an uncertainty head for confidence estimation. The masking-based training further enables 4D prediction including interpolation, forecasting, and backcasting. To address limited 4D supervision, we initialize from large-scale 3D rigging data and then adapt to mid-scale 4D rigging data. Experiments show that \ourmethod achieves strong static 3D rigging performance with controllable bone complexity, as well as high-quality, temporally consistent 4D reconstruction and prediction.

Our main contributions are as follows:
\begin{itemize} [leftmargin=*, topsep=0pt, itemsep=0pt]
  \item We propose \ourmethod, a unified flow-matching model for controllable 3D/4D skeleton generation.
  \item We introduce a 3D-to-4D curriculum for data-efficient training under limited 4D supervision.
  \item We achieve strong 3D rigging, high-quality 4D reconstruction, and flexible 4D prediction.
\end{itemize}

\section{Related Work}
\textbf{Static Automatic Rigging and Skinning.}
Automatic rigging has been studied extensively in the static 3D setting. Early methods relied on template fitting and per-shape optimization, as in Pinocchio~\citep{baran2007automatic}, while later data-driven approaches improved both speed and generalization. RigNet~\citep{xu2020rignet} introduced graph neural networks for direct skeleton and skinning prediction from meshes. Subsequent methods such as UniRig~\citep{zhang2025one}, RigAnything~\citep{liu2025riganything}, ARMO~\citep{sun2025armo}, Puppeteer~\citep{song2025puppeteer}, and Skin Tokens~\citep{zhang2026skin} further improved structural validity through autoregressive or language-model-style generation~\citep{song2025magicarticulate}. These methods are strong baselines for static 3D rigging, but not for general-purpose 4D rigging. In particular, autoregressive generation is less suitable for 4D sequences, where one must reason jointly over skeletal slots and time; applying static models per-frame often leads to temporal flickering and inconsistent topologies~\citep{wang2026sprig}. Their discrete formulation can also make skeleton density control harder for downstream 4D use.

\textbf{Dynamic and 4D Rigging.}
As 4D meshes become more accessible, there is growing interest in rigging dynamic assets directly from videos or mesh sequences. Existing 4D rigging-related methods mainly fall into two categories. The first consists of per-sequence optimization methods, such as AKD~\citep{li2025articulated}, RigGS~\citep{yao2025riggs}, AniMimic~\citep{xie2026animimic}, GaussiAnimate~\citep{wang2026gaussianimate}, and RAT4D~\citep{dabhi2026rat4d}, which align articulated structure to multi-frame observations or dynamic 3D representations. These methods can achieve high fidelity, but their reliance on per-scene optimization makes them computationally expensive and less suitable for scalable deployment. The second involves consistency-oriented extensions of static riggers, such as SPRig~\citep{wang2026sprig}, which improve cross-frame consistency through additional self-supervision. While promising, these methods do not yet provide a unified feedforward framework for raw 4D mesh sequences that supports both reconstruction and prediction. Related animation methods, including Gaussian bone representation, zero-shot pose transfer and morphology-aware motion transfer~\citep{guo2025makeitposeable, muralikrishnan2025smf, kim2026morphgs, zhang2026rigmo}, are complementary rather than direct solutions to general-purpose 4D rigging.

\textbf{4D Mesh Availability and Upstream Geometry Models.}
Our work is motivated in part by the increasing availability of temporally coherent 4D meshes from upstream reconstruction and generation systems. The field has seen rapid expansion, ranging from text- and image-to-4D generation~\citep{yin20234dgen, Ling_2024_CVPR, bahmani2024tc4d, miao2024pla4d, yuan20244dynamic, chen2024ct4d, zeng2024trans4d, Bahmani_2024_CVPR, sun2025eg4d, lin2025phy124, lin2025phys4dgen, zhao2023animate124, gao2024gaussianflow, wang2024animatabledreamer, dai2025textmesh4d, nag20252} to video-based dynamic reconstruction~\citep{jiang2024consistent4d, Zeng_2024_ECCV, wu2024sc4d, zhang2024magicpose4d, zhang20244diffusion, li2024dreammesh4d, fu2024sync4d, zhu2026ar4d, yao2025sv4d, xie2025sv4d, chen2025v2m4, ren2024l4gm, ren2023dreamgaussian4d, liang2024diffusion4d}. In particular, recent methods such as ActionMesh~\citep{sabathier2026actionmesh} and AniMimic~\citep{xie2026animimic} show that high-quality dynamic geometry can now be derived from monocular videos and learned priors. While these systems expand the supply of dynamic geometry, they do not solve the downstream rigging problem. Instead, they highlight the need for a feedforward 4D rigging model that converts temporally aligned geometry into informative and animation-ready skeletons. In this sense, our method is complementary: upstream models provide the dynamic geometry, while \ourmethod provides the rigging for downstream animation and control.

\mypara{Generative Models for 3D and 4D Assets.}
A separate line of work studies the co-generation of geometry and rigging information. AniGen~\citep{huang2026anigen} generates animatable 3D assets together with rigging information, while AKD~\citep{li2025articulated} uses score distillation to generate articulated 4D assets from images or text. Related directions include skeleton-guided 3D generation~\citep{xu2025skdream}, creative articulated asset generation~\citep{lu2026muses}, sketch-based rigging~\citep{zhao2026stroke3d}, and rig-based generative synthesis of dynamic 3D Gaussians~\citep{wu2026skeletongaussian}. However, these works are not for general-purpose feedforward rigging of raw 4D mesh sequences. Instead, our focus is to take 3D or 4D geometry as input and produce temporally consistent skeletons with controllable complexity and prediction capability, making our setting complementary to asset generation.

\section{Proposed Method}
% \subsection{One-to-one flow-matching skeleton generation for dynamic meshes}

\begin{figure}[t]
\centering
\includegraphics[width=\linewidth]{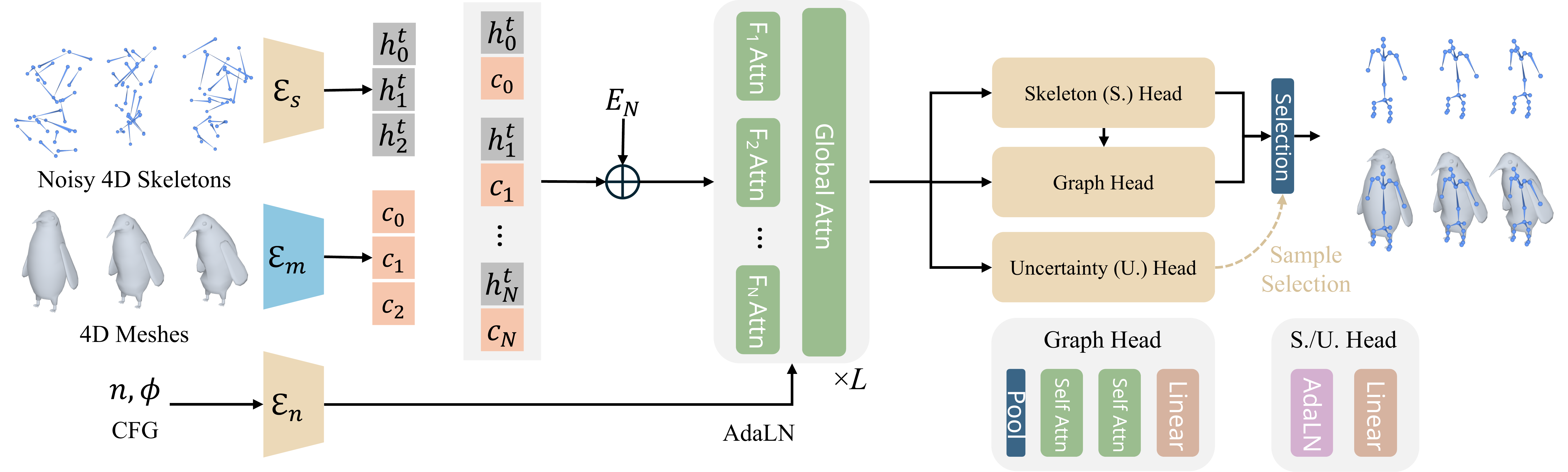}
\vspace{-6mm}
\caption{\small \ourmethod overview. We encode meshes into conditioning tokens, and initialize noisy skeleton slots, which are processed by a flow matching transformer to predict temporally aligned skeleton representations. The predicted features are decoded by the skeleton, graph, and uncertainty heads to recover 3D/4D rigging outputs, with uncertainty used for sample selection. $n$ denotes the skeleton count, and $\phi$ denotes the unconditional token.}
\label{fig:framework}
\end{figure}
% \vspace{-10mm}

\subsection{Problem Formulation}
Let $\mathcal{M}_{1:N}=(m_1,\dots,m_N)$ denote a observation sequence of mesh $m_i$, where each frame is represented as a sampled point cloud with normals $P_i\in\mathbb{R}^{N_{p}\times 6}$ and $N_p$ is the size of the point cloud. Our goal is to predict a corresponding skeleton sequence $\mathcal{S}_{1:N}=(S_1,\ldots,S_N)$, where each $S_i$ is a set of bones with 3D endpoints and a tree topology. We consider static and dynamic tasks below.

\textbf{Static skeleton reconstruction.}
When $N=1$, we generate a skeleton set $S_1$ conditioned on a single mesh observation $P_1$. Formally, we model the conditional distribution $p(S_1\mid P_1)$.

\textbf{Dynamic skeleton reconstruction and prediction.}
Given observed frames $\mathcal{O}\subseteq\{1,\ldots,N\}$ and observations $\mathcal{P}_{\mathcal{O}}=\{P_i\}_{i\in\mathcal{O}}$, we generate the full skeleton sequence using $p(\mathcal{S}_{1:N}\mid \mathcal{P}_{\mathcal{O}})$. Reconstruction corresponds to $\mathcal{O}=\{1,\ldots,N\}$, while prediction corresponds to $\mathcal{O}\subset\{1,\ldots,N\}$.

\subsection{Flow Matching on An Optimal-Transport Path}
Instead of learning a reverse denoising process as in diffusion models, flow matching learns a continuous-time vector field that transports noise to data. Let $\mathbf{x}_0$ denote a clean sample and $\mathbf{x}_1\sim\mathcal{N}(0,\sigma^2 I)$ be Gaussian noise. We use the optimal-transport straight-line path $\mathbf{x}_t=(1-t)\mathbf{x}_0+t\mathbf{x}_1$ for $t\in(0,1)$, whose target velocity is constant: $\mathbf{v}^{\star}(\mathbf{x}_t,t)=\mathbf{x}_1-\mathbf{x}_0$.
\
Given mesh-derived conditioning tokens $\mathbf{c}$, the network $\mathcal{M}$ predicts the velocity field $\hat{\mathbf{v}}=\mathcal{M}(\mathbf{x}_t,t,\mathbf{c};\Theta)$ and is trained to match $\mathbf{v}^{\star}$ with mean-squared error, together with our matching and auxiliary losses.
\
At inference time, we start from $\mathbf{x}_{t=1}\sim\mathcal{N}(0,\sigma^2 I)$ and integrate the learned ODE from $t=1$ to $t=0$ with an Euler solver:
$\mathbf{x}_{t-\Delta t}=\mathbf{x}_t-\Delta t\cdot \hat{\mathbf{v}}(\mathbf{x}_t,t,\mathbf{c})$, yielding the clean sample $\mathbf{x}_{0}$.

\subsection{4D Skeleton Flow Matching Model}

Figure~\ref{fig:framework} shows an overview of \ourmethod.
Given a static mesh or a mesh sequence, we encode each frame with the frozen Michelangelo encoder~\citep{zhao2023michelangelo} to obtain mesh conditioning tokens $\mathbf{c}\in\mathbb{R}^{N\times L_m\times H}$. In parallel, we initialize a fixed set of noisy skeleton slots $\mathbf{x}_t\in\mathbb{R}^{N\times L_s\times 7}$ and optionally inject a target skeleton-count condition to control the number of generated bones. We set $L_s=128$ to cover all training instances, similar to prior slot-based methods such as Puppeteer~\citep{song2025puppeteer}, and allow $N$ to vary by sampling different window lengths during training. The DiT embeds the skeleton slots into $H$-dimensional tokens, concatenates them with the mesh tokens, and processes the resulting sequence of length $N\cdot(L_s+L_m)$ with temporal rotational positional encoding (RoPE) using the frame index as the shared temporal reference. It then performs slot-based set prediction over the full sequence and predicts a velocity tensor $\hat{\mathbf{v}}\in\mathbb{R}^{N\times L_s\times 7}$, which is decoded into up to $L_s$ bones per frame as 7D slot states $\mathbf{x}_0\in\mathbb{R}^{N\times L_s\times 7}$, where each slot is $\mathbf{x}_{0,i,j}=[e_{i,j},\mathbf{p}_{i,j,0},\mathbf{p}_{i,j,1}]$ with an existence indicator $e\in\{0,1\}$ and two 3D endpoints $\mathbf{p}_{i,j,0},\mathbf{p}_{i,j,1}\in\mathbb{R}^3$. After the transformer predicts the aligned skeleton slots, a graph head infers the kinematic tree and an uncertainty head estimates confidence, producing the final controllable 3D or 4D rigging output.

Besides the skeleton head, we use two additional lightweight heads: a graph head and an uncertainty head. A naive way to build a skeleton tree is to connect bones by distance, but this can lead to unstable topology over time as the joints move. Instead, the graph head predicts parent indices from the hidden features $\{\mathbf{h}_i\}$ and the predicted skeletons $\{\hat{\mathbf{x}}_i\}$, yielding a more temporally consistent tree structure. The uncertainty head takes the hidden features as input and predicts per-frame per-bone uncertainty $\hat{\sigma}\in\mathbb{R}^{N\times L_s\times 7}$. Intuitively, it estimates how reliable each generated bone is, so we can use uncertainty as a score when drawing multiple samples and select the most confident result.

\textbf{Skeleton-count conditioning and classifier-free guidance (CFG).}
In addition to mesh tokens, we condition the model on a scalar skeleton or bone count $y$, which serves as a guidance label. During training we apply classifier-free guidance by randomly dropping the condition with probability $p_\textrm{drop}$, that is, replacing $y$ with an unconditional token $y=-1$ in the embedding layer. The conditional and unconditional predictors share parameters and differ only in the presence of $y$:
\begin{equation}
\hat{\mathbf{v}}_\textrm{cond}=\mathcal{M}(\mathbf{x}_t,t,\mathbf{c},y;\Theta), \qquad
\hat{\mathbf{v}}_\textrm{uncond}=\mathcal{M}(\mathbf{x}_t,t,\mathbf{c},-1;\Theta).
\end{equation}
During inference, we form the guided velocity with CFG scale $s$ as $\hat{\mathbf{v}}_\textrm{cfg}=\hat{\mathbf{v}}_\textrm{uncond}+s\cdot\big(\hat{\mathbf{v}}_\textrm{cond}-\hat{\mathbf{v}}_\textrm{uncond}\big)$, and integrate the ODE with $\hat{\mathbf{v}}_\textrm{cfg}$.

\begin{figure}[t]
\centering
\includegraphics[width=1\linewidth]{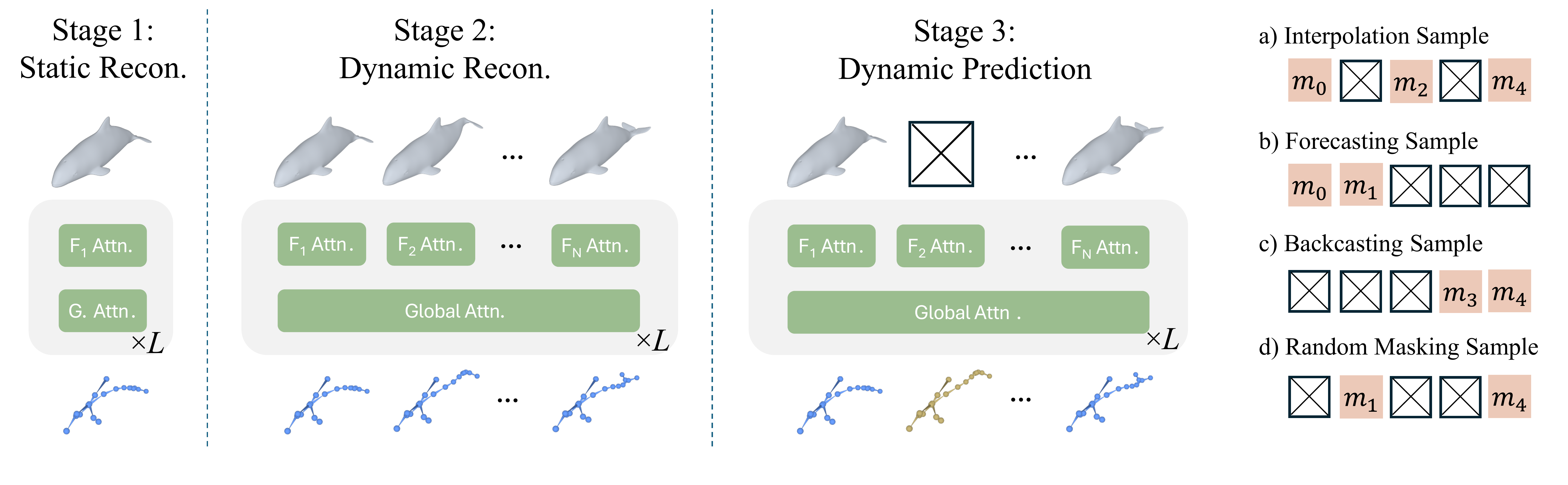}
\vspace{-10mm}
% \fbox{\rule{0pt}{2in}\rule{0.9\linewidth}{0pt}}
\caption{\small Data-efficient curriculum training. We first train on static 3D rigging, then adapt to 4D reconstruction, and finally enable prediction through masked conditioning. During the final stage, we use diverse masking patterns, including interpolation, forecasting, backcasting, and random masking, to expose the model to different temporal prediction settings. This schedule transfers large-scale 3D supervision to the data-scarce 4D setting.}
\label{fig:curriculum}
\end{figure}

\subsection{Training Objective}
Our training objective combines flow matching with the auxiliary supervision terms.

\textbf{One-to-one supervision via sequence Hungarian matching.}
To handle variable ground-truth bone counts while preserving a fixed tensor interface, we follow the DETR-style one-to-one supervision \citep{carion2020end}. Each $S_i$ is represented using $L_s$ permutation-invariant skeleton slots, and the model is trained with hungarian matching for each predicted and ground-truth skeleton pair. For each frame we extract raw endpoints $\mathbf{g}_i\in\mathbb{R}^{N_\textrm{gt}\times 6}$ and pad them to a fixed tensor size. Since slot indices are permutation-invariant, we compute a single assignment per sequence, shared across frames, using the Hungarian algorithm. We first recover a predicted clean state by
\begin{align}
\hat{\mathbf{x}}_0=\mathbf{x}_t-t\,\hat{\mathbf{v}}\in\mathbb{R}^{N\times L_s\times 7},    
\end{align}
and build a cost matrix between slot endpoints $\hat{\mathbf{p}}_{j,0}$ and $\hat{\mathbf{p}}_{j,1}$ and the ground-truth endpoints $\mathbf{g}_j$. 
To align the static and dynamic training objectives, we set the anchor frame $i=1$ for hungarian matching in all training stages. The matching cost is a weighted sum of endpoint Euclidean distance. The resulting assignment maps each ground-truth bone to one slot, producing matched targets $\tilde{\mathbf{x}}$.

\textbf{Flow matching loss.}
The primary objective is a weighted flow-matching mean-squared error on the matched optimal-transport velocity:
\begin{align}
\mathcal{L}_\textrm{FM}=\frac{1}{N L_s}\sum_{i,j} w_{i,j}\,\left\lVert \hat{\mathbf{v}}_{i,j}-\tilde{\mathbf{v}}^{\star}_{i,j}\right\rVert_2^2,
\qquad
\tilde{\mathbf{v}}^{\star}=\mathbf{x}_1-\tilde{\mathbf{x}}_0.
\end{align}
where per-slot weights $w\in\mathbb{R}^{N\times L_s}$ upweight matched slots and control supervision on empty slots with less focus. Intuitively, we set the unmatched slot endpoints with zero weight since they have no contribution to the final results, thus forcing the model to focus on the matched pairs.

\textbf{Uncertainty loss.}
% When the DiT predicts per-channel uncertainty $\hat{\boldsymbol{\sigma}}_v$,
To quantify the confidence of skeletons, we add a Gaussian negative log-likelihood (NLL) term to estimate the uncertainty of bone existence and the joint positions. Following IDDPM~\citep{nichol2021improved}, we stop gradients through the mean prediction in the NLL branch:
\begin{equation}
\mathcal{L}_\textrm{NLL}=\frac{1}{N L_s}\sum_{i,j} w_{b,j}
\Big(
{\left\lVert \mathrm{sg}(\hat{\mathbf{v}}_{b,i,j})-\tilde{\mathbf{v}}^{\star}_{b,i,j}\right\rVert_2^2} / ({\hat{\boldsymbol{\sigma}}_{v,b,i,j}^2+\epsilon})
+\log\big(\hat{\boldsymbol{\sigma}}_{v,b,i,j}^2+\epsilon\big)
\Big),
\end{equation}
where $\epsilon$ is a small value and $\mathrm{sg}(\cdot)$ denotes the stop-gradient operator. 
This way, $\mathcal{L}_{NLL}$ can guide $\hat{\sigma}^2$ without affecting the training of transformer backbone and skeleton head. 

\textbf{Graph head loss.}
To build the kinematic tree for the generated skeleton, we supervise a parent index classifier over slots. Specifically, to handle temporal aggregation and derive a temporal-consistent tree, we pool time by averaging to obtain a single parent distribution per slot for geometry $\bar{\mathbf{x}}_0=\frac{1}{N}\sum_{i=1}^{N}\hat{\mathbf{x}}_{0,i}$ and high-level context $\bar{\mathbf{h}}=\frac{1}{N}\sum_{i=1}^{N}\mathbf{h}_i$, concatenating them to form the input of the graph head $\mathbf{z}=\mathrm{concat}\big(\mathrm{sg}(\bar{\mathbf{h}}),\mathrm{sg}(\bar{\mathbf{x}}_0)\big)\in\mathbb{R}^{L_s\times (H+7)}$. The graph head $G$ will output $\mathbf{p}=G(\mathbf{z})\in\mathbb{R}^{L_s\times (L_s+2)}$ where the additional two channels are used to represent the root and non-existing classes.
Targets are constructed from the ground-truth joint graph and the Hungarian assignment, including root, non-root, and non-existing classes. The loss is
\begin{equation}
\mathcal{L}_\textrm{graph}=\frac{1}{L_s}\sum_j \mathrm{CE}(\mathbf{p}_{b,j,:}, y^\textrm{parent}_{b,j})\cdot (1-t)^{\gamma_p},    
\end{equation}
where $\gamma_p$ is a hyperparameter to control the denosing-step-aware weight $(1-t)^{\gamma_p}$ in flow matching. Intuitively, small $t$ means the noisy state is still close to a coherent skeleton, so parent index targets (via hungarian matching on $\hat{\mathbf{x}}_0$) are more reliable while large $t$ means slots are dominated by noise, so downweighting parent CE reduces conflicting topology gradients. The exponent $\gamma_p\ge 0$ controls how sharply supervision is concentrated toward $t\approx 0$.

\mypara{Total objective.}
The final objective is a weighted sum:
\begin{equation}
\mathcal{L}=\mathcal{L}_\textrm{FM}+\lambda_\textrm{NLL}\mathcal{L}_\textrm{NLL}+\lambda_\textrm{graph}\mathcal{L}_\textrm{graph}+\lambda_\textrm{aux}\mathcal{L}_\textrm{aux},    
\end{equation}
where $\lambda$ are the coefficients for each term and $\mathcal{L}_\textrm{aux}$ is used to regularize the geometry and temporal smoothness (defined in the Appendix~\ref{app:train_detail}). Overall, the losses in our framework serve three purposes: (1) $\mathcal{L}_\textrm{FM}$ learns accurate skeleton generation through flow matching, (2) $\mathcal{L}_\textrm{NLL}$ and $\mathcal{L}_\textrm{graph}$ improve confidence estimation and kinematic topology prediction, and (3) $\mathcal{L}_\textrm{aux}$ regularizes geometric validity and temporal consistency. Together, these terms improve stability and produce controllable, temporally aligned skeletons for both 3D and 4D inputs.

\subsection{Curriculum Training from 3D Reconstruction to 4D Prediction}
We train the model with a curriculum progressively increasing temporal complexity while keeping the architecture fixed. As shown in Figure~\ref{fig:curriculum}, we start with static single-frame skeleton generation initializing the model with a strong geometry prior. We then switch to dynamic reconstruction, training on different time windows with full guidance of mesh. Finally, we emphasize dynamic prediction by randomly masking mesh tokens to force the model to simulate missing observations.

\section{Experiments}
We evaluate \ourmethod across diverse rigging scenarios, focusing on its ability to maintain 4D consistency and performance as a static rigger. Unless otherwise specified, we use unconditional guidance at inference time, since the ground-truth bone count is unavailable at test time.

\subsection{Experimental Setup}
\label{subsec:setup}

\textbf{Datasets.} 
To ensure a comprehensive evaluation, we utilize three primary data sources spanning static and dynamic domains. (1) For static rigging, we follow the protocol established in Puppeteer \citep{song2025puppeteer} and employ Articulation-XL \citep{song2025magicarticulate}, a massive collection of 33,000 high-quality rigged 3D models covering humanoids, animals, and mechanical objects. (2) We also evaluate on the classic ModelsResource dataset \citep{xu2020rignet}, which contains a curated set of character models with professionally designed skeletons, serving as a standard benchmark for topological accuracy. (3) For dynamic 4D rigging, we utilize a curated dynamic dataset derived from TexVerse~\citep{zhang2025texverse}. We manually filtered the TexVerse animated subset to retain around 2,500 object-centric instances where the skeleton and mesh remain strictly aligned across temporal sequences, providing a high-fidelity ground truth for evaluating consistency and predictive forecasting. More details about the curated dataset can be found in Appendix~\ref{app:curated_dataset}.

\textbf{Evaluation metrics.} We evaluate skeleton quality with three Chamfer Distance (CD) variants: Joint-to-Joint (J2J), Joint-to-Bone (J2B), and Bone-to-Bone (B2B), which measure joint accuracy, joint-to-bone alignment, and structural fidelity, respectively. For 4D tasks, we average these over time. We also evaluate temporal consistency using bone distance deviation (BDD), which measures temporal jitter in bone lengths; Graph Structural Distance (GSD), which measures structural consistency through Hamming distance (HD)~\citep{wang2026sprig}; and bone number deviation (BND), which measures changes in bone count relative to the anchor frame. More details are provided in Appendix~\ref{app:evaluation_metrics}.

\textbf{Baselines.} We compare \ourmethod against both classical and modern rigging baselines. Classical methods include Pinocchio~\citep{baran2007automatic}, a template-fitting optimization-based rigging system, and RigNet~\citep{xu2020rignet}, a graph-neural-network-based feedforward rigger. Among recent generative or transformer-based methods, we consider MagicArticulate~\citep{song2025magicarticulate}, Anymate~\citep{deng2025anymate}, UniRig~\citep{zhang2025one}, RigAnything~\citep{liu2025riganything}, and Puppeteer~\citep{song2025puppeteer}. These methods cover a broad spectrum from optimization-based pipelines to modern feedforward sequence models. While several of them are strong for static 3D meshes, our results show that \ourmethod remains competitive in the static setting while additionally providing the native 4D consistency and predictive capability required for dynamic sequences.

\begin{table}[t]
\small
\centering\tabcolsep 6.5pt % reduce horizontal spacing between columns
\caption{\small \textbf{Static skeleton reconstruction comparison.} Our methods achieves strong reconstruction fidelity across diverse out-of-domain static datasets. \textit{Ours*} utilizes uncertainty as selection score to select the best candidate from the 5 samples and outperforms all baselines on Articulation-XL 2.0 and Diverse-Pose. On ModelsResource, its zero-shot performance approaches that of RigAnything trained directly on the same dataset. See the "Baselines" paragraph for citations of these models.}
\vspace{-2mm}
\label{tab:static_reconstruction_comparison}
\renewcommand{\arraystretch}{1.1}
% \resizebox{\textwidth}{!}{%
\begin{tabular}{l|ccc|ccc|ccc}
\toprule
& \multicolumn{3}{c|}{Articulation-XL 2.0} & \multicolumn{3}{c|}{ModelsResource} & \multicolumn{3}{c}{Diverse-Pose} \\
Model & J2J $\downarrow$ & J2B $\downarrow$ & B2B $\downarrow$ & J2J $\downarrow$ & J2B $\downarrow$ & B2B $\downarrow$ & J2J $\downarrow$ & J2B $\downarrow$ & B2B $\downarrow$ \\
\midrule
Pinocchio & 0.0832 & 0.0661 & 0.0549 & 0.0685 & 0.0482 & 0.0409 & 0.0797 & 0.0641 & 0.0515 \\
RigNet & 0.0762 & 0.0608 & 0.0528 & 0.0722 & 0.0599 & 0.0433 & 0.0775 & 0.0639 & 0.0571 \\
MagicArt. & 0.0326 & 0.0250 & 0.0212 & 0.0411 & 0.0314 & 0.0269 & 0.0438 & 0.0346 & 0.0296 \\
Anymate & 0.0409 & 0.0301 & 0.0294 & 0.0370 & 0.0261 & 0.0258 & 0.0342 & 0.0255 & 0.0272 \\
UniRig & 0.0331 & 0.0261 & 0.0218 & 0.0396 & 0.0302 & 0.0257 & 0.0325 & 0.0257 & 0.0208 \\
RigAnyth. & 0.0489 & 0.0379 & 0.0306 & \textbf{0.0343} & \textbf{0.0244} & \textbf{0.0204} & 0.0397 & 0.0308 & 0.0241 \\
Puppeteer & 0.0311 & 0.0237 & 0.0198 & 0.0377 & 0.0280 & 0.0241 & \underline{0.0251} & 0.0199 & \underline{0.0160} \\
Ours & \underline{0.0264} & \underline{0.0188} & \underline{0.0163} & {0.0358} & {0.0259} & {0.0234} & 0.0252 & \underline{0.0188} & \underline{0.0160} \\
Ours* & \textbf{0.0259} & \textbf{0.0184} & \textbf{0.0160} & \underline{0.0354} & \underline{0.0256} & \underline{0.0233} & \textbf{0.0245} & \textbf{0.0182} & \textbf{0.0155} \\
\bottomrule
\end{tabular}%
% }
\end{table}
% \vspace{-15mm}

\begin{figure}[t]
\centering
\includegraphics[width=\linewidth]{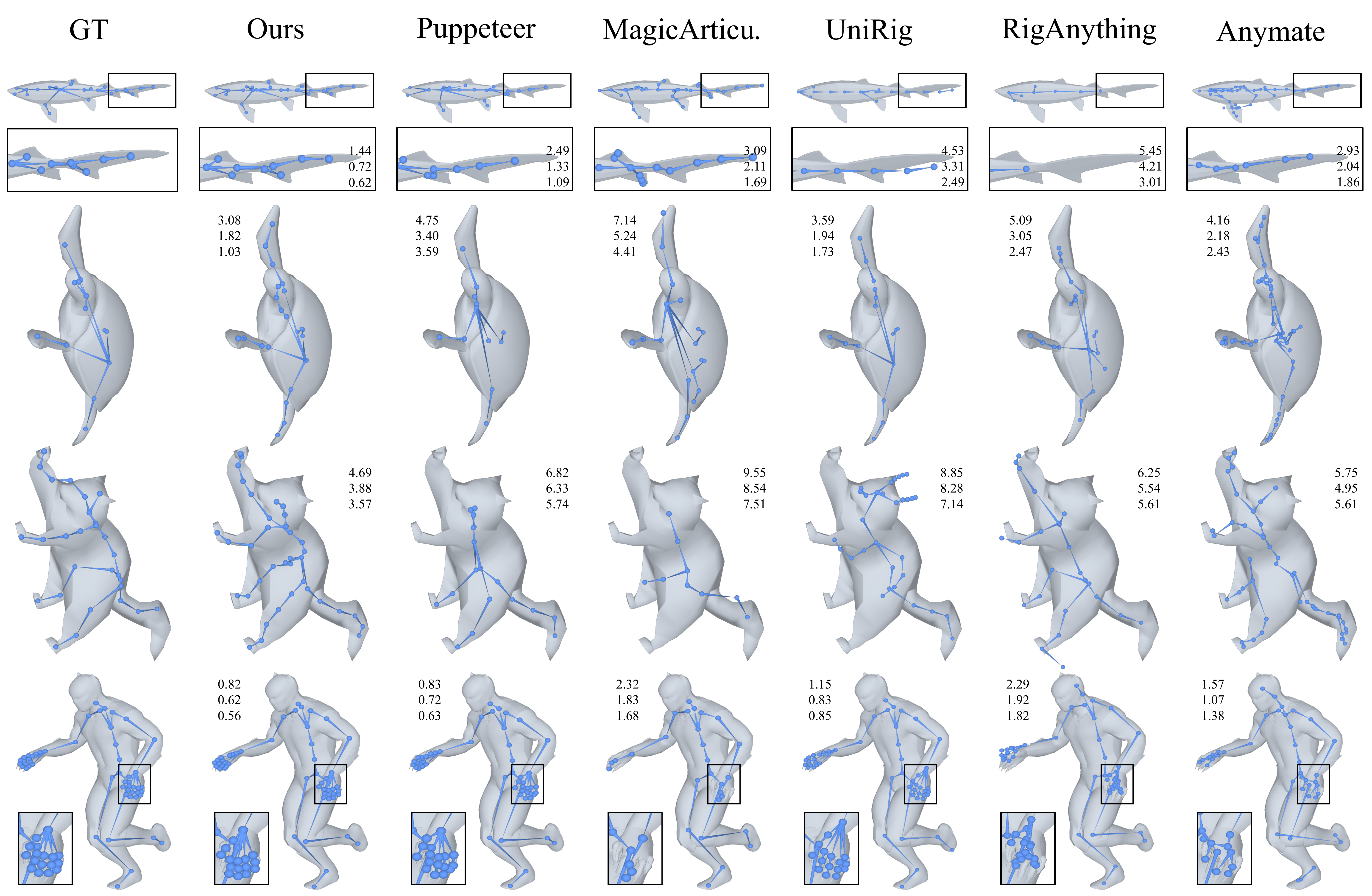}
\caption{\small Static Reconstruction. Numbers from top to bottom are J2J, J2B, and B2B for each instance. Compared with baselines, our model preserves finer geometric details and achieves more accurate mesh alignment.}
\label{fig:static}
\end{figure}

\subsection{3D Skeleton Reconstruction}
\ourmethod supports both static and dynamic meshes, and with a single mesh as input it becomes a strong 3D rigging model. As shown in Table~\ref{tab:static_reconstruction_comparison}, our method ranks among the best across all three datasets, clearly outperforming classical baselines such as Pinocchio and RigNet. Among recent methods, Puppeteer~\citep{song2025puppeteer} and RigAnything~\citep{liu2025riganything} are the strongest competitors, but \ourmethod achieves the best overall results on Articulation-XL 2.0 and Diverse-Pose and remains competitive on ModelsResource in the zero-shot setting, even if RigAnything is trained directly on that training dataset. Overall, these results show that our unified 3D/4D formulation preserves strong static rigging performance and improves further with uncertainty-based selection in \textit{Ours*}. Figure~\ref{fig:static} shows that our method recovers more detailed and better mesh-aligned skeletons compared with baselines; more results are shown in Appendix~\ref{app:more_results} and the demo video.

\begin{figure}[t]
\centering
\begin{subfigure}[t]{0.49\linewidth}
    \centering
    \includegraphics[width=\linewidth]{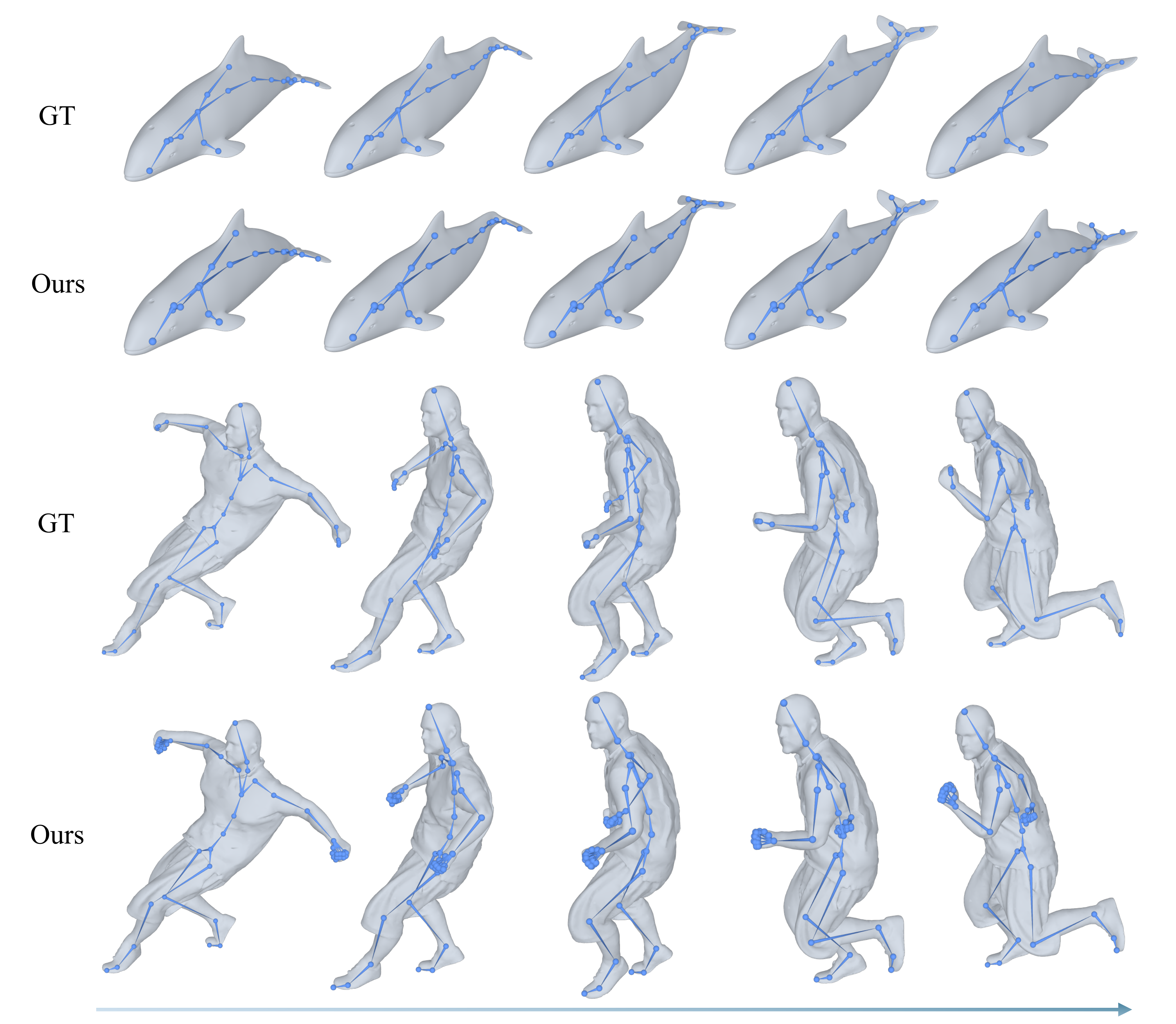}
    \caption{\small Dynamic Reconstruction}
    \label{fig:dy_recon}
\end{subfigure}\hfill
\begin{subfigure}[t]{0.49\linewidth}
    \centering
    \includegraphics[width=\linewidth]{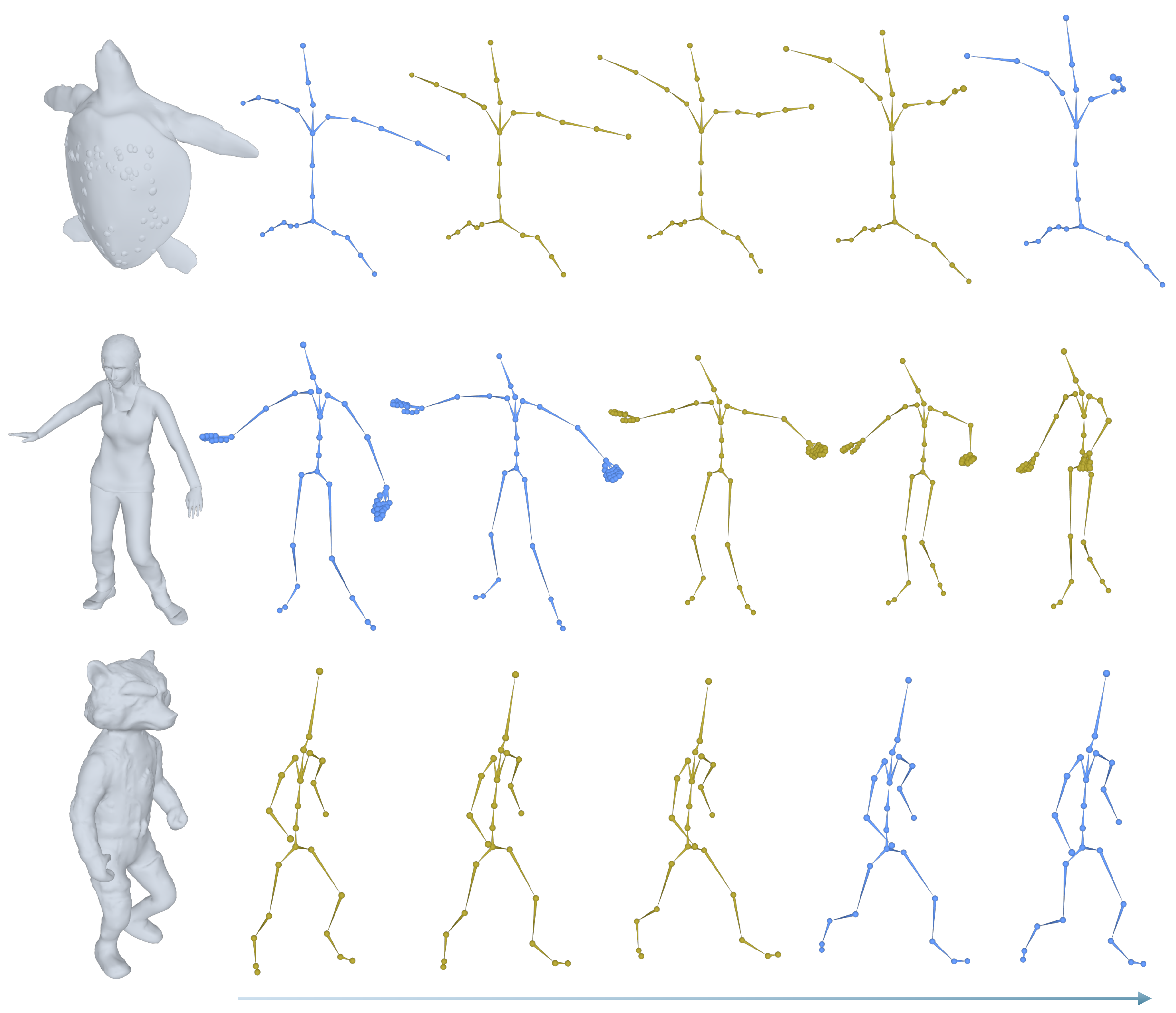}
    \caption{\small Dynamic prediction.}
    \label{fig:dy_pred}
\end{subfigure}
\vspace{-3mm}
\caption{\small Dynamic reconstruction and prediction. Blue indicates reconstruction with mesh guidance, while yellow indicates predictions without mesh guidance. In (b), from top to bottom, it shows interpolation, forecasting, and backcasting, respectively. Refer to the demo video for a clearer comparison.}
\end{figure}

\begin{table}[t]
\small
\tabcolsep 7pt %% control the table width
\centering
\caption{\small Dynamic quantitative comparisons on TexVerse dynamic dataset. $N$ is time window length. \textit{Puppeteer-D*} means we finetuned Puppeteer-diverse model~\citep{song2025puppeteer} on TexVerse training dataset. \textit{Ours*} utilizes uncertainty as selection score to select the best candidate from the 5 samples.}
\vspace{-2mm}
\label{tab:dynamic_reconstruction}
\begin{tabular}{cc|ccc|cccc}
\toprule
Method & $N$ & J2J-W $\downarrow$ & J2B-W $\downarrow$ & B2B-W $\downarrow$ & BND $\downarrow$ & BDD $\downarrow$ & GSD $\downarrow$ & HD $\downarrow$ \\
\midrule
Puppeteer-D* & 4 &  0.0161 &	0.0130	&0.0118   &  2.8279	& 0.0075	& 0.1189	& 6.2173   \\
Puppeteer-D* & 8 &  0.0162 &	0.0131 &	0.0118   &    2.8753 &	0.0091	& 0.1575	& 8.3745     \\
Puppeteer-D* & 24 & 0.0163    & 0.0132    & 0.0119    &     3.1763	& 0.0150	& 0.2654	& 14.1894  \\
\midrule
Ours &4 & 0.0161 & 0.0125 & 0.0104 & 0 & 0.0021 & 0 & 0 \\
Ours & 8 & 0.0152 & 0.0119 & 0.0099 & 0 & 0.0026 &  0 & 0 \\
Ours & 24 & 0.0142 & 0.0112 & 0.0093 & 0 & 0.0038    &  0 & 0 \\
\midrule
Ours* & 4 & 0.0144 &	0.0112 &	0.0093 & 0 & 0.0019 & 0 & 0\\
Ours* & 8 & 0.0144	& 0.0113	& 0.0094 & 0 & 0.0025 & 0 & 0\\ 
Ours* & 24 & 0.0136 & 0.0107 & 0.0089 & 0 & 0.0037  & 0 & 0 \\
\bottomrule
\end{tabular}
\end{table}

\subsection{4D Skeleton Reconstruction and Prediction}
Our model naturally handles 4D reconstruction and prediction. For dynamic meshes, we study two settings. In 4D reconstruction, the full mesh sequence is observed with the goal of recovering a temporally consistent skeleton for every frame. In 4D prediction, only part of the sequence is observed and the model must infer the missing skeletons for interpolation, forecasting, or backcasting. Table~\ref{tab:dynamic_reconstruction} shows that \ourmethod consistently outperforms the fine-tuned Puppeteer baseline \textit{Puppeteer-D*} on both reconstruction accuracy and 4D consistency. As the temporal window $N$ increases, our model achieves lower J2J-W, J2B-W, and B2B-W errors, where W represents averaging over the time window, indicating more accurate per-frame skeleton reconstruction with stronger temporal context. More importantly, \ourmethod maintains perfect structural consistency across the full sequence, with zero BND, GSD, and HD for all reported settings, meaning that the predicted bone count and kinematic topology remain stable over time. In contrast, \textit{Puppeteer-D*} exhibits clear temporal degradation as $N$ grows, with increasing BND, BDD, GSD, and HD, indicating both geometric drift and topological flickering. The uncertainty-based selection in \textit{Ours*} further improves reconstruction quality while preserving the same structural behavior, showing that \ourmethod produces not only accurate per-frame skeletons but also temporally coherent 4D rigs. Figure~\ref{fig:dy_recon} provides qualitative examples showing that our model preserves 4D consistency over time while also recovering finer skeletal details, such as the structure of the human hand.
In addition, \ourmethod supports 4D prediction, enabling interpolation, forecasting, and backcasting within a unified framework. Figure~\ref{fig:dy_pred} presents qualitative evaluations of these settings and demonstrates \ourmethod's flexibility for downstream tasks. More qualitative results are shown in Appendix~\ref{app:more_results} and the demo video.

\subsection{Further Analysis and Ablation Studies}

% \begin{wraptable}{r}{0.5\linewidth}
% \vspace{-0.8em}
\begin{table}[t]
\small
\tabcolsep 10pt %% control the table width
\centering
\caption{\small Ablation studies on data sources. Metrics are evaluated on the TexVerse test set with $T=24$.} 
\vspace{-2mm}
\label{tab:ab_dynamic_reconstruction}
% \resizebox{\linewidth}{!}{%
\begin{tabular}{cc|ccc|cc}
\toprule
Static & Dynamic & J2J-W $\downarrow$ & J2B-W $\downarrow$ & B2B-W $\downarrow$ & BDD $\downarrow$ & GSD $\downarrow$ \\
\midrule
\xmark & \cmark & 0.0213 & 0.0173 & 0.0147 & 0.0049 & 0\\
\cmark & \cmark & \textbf{0.0142} & \textbf{0.0112} & \textbf{0.0093} & \textbf{0.0038} & 0\\
\bottomrule
\end{tabular}%
% \vspace{-1mm}
\end{table}
\vspace{-3mm}
% }

% \vspace{-1.0em}
% \end{wraptable}
\textbf{Efficient Training Curriculum.} We apply curriculum training: pre-train on large-scale static assets and fine-tune on dynamic sequences. Table~\ref{tab:ab_dynamic_reconstruction} shows that static pre-training improves accuracy and reduces temporal instability compared with training on dynamic data alone. This suggests that large-scale static supervision provides a strong geometric prior for learning skeletal dynamics.

\begin{wraptable}{r}{0.3\linewidth}
\vspace{-0.8em}
\small
\tabcolsep 2pt %% control the table width
\centering
\vspace{-1.5mm}
\caption{\small Ablation studies on graph head. The metrics are evaluated on the TexVerse test set with time window $T=24$.}
\vspace{-2mm}
\label{tab:ab_graph_head}
\begin{tabular}{c|cc}
% \vspace{-2mm}
\toprule
Graph Head & GSD $\downarrow$ & HD $\downarrow$ \\
\midrule
\xmark & 0.0375 & 2.2844 \\
\cmark & 0 & 0 \\
\bottomrule
\end{tabular}
\vspace{-1.0em}
\end{wraptable}

\textbf{Ablation of the Graph Head.} The graph head is introduced to maintain a temporally consistent kinematic hierarchy across the full 4D sequence. We evaluate its effect by replacing it with a simple 1-Nearest Neighbor tree-building heuristic. As shown in Table~\ref{tab:ab_graph_head}, removing the graph head significantly increases Hamming distance, indicating frequent topological changes and unstable parent assignments over time. The corresponding rise in GSD further shows that the graph head is essential for preserving structural coherence and avoiding topological flickering in dynamic rigging.

\begin{wrapfigure}{r}{0.46\linewidth}
\vspace{-1.2em}
\centering
\includegraphics[width=\linewidth]{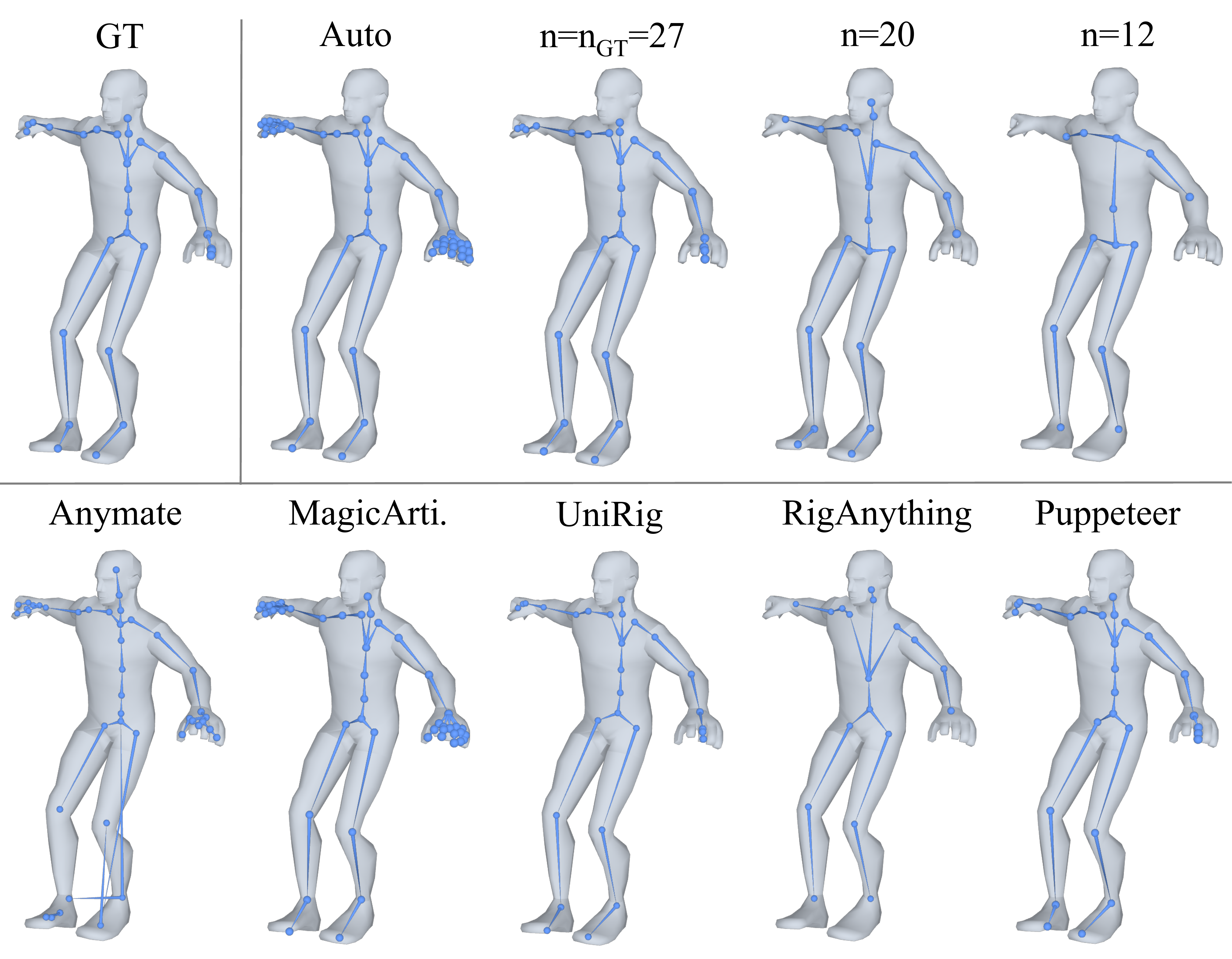}
\caption{\small Analysis of bone count control. Our model can freely control the number of bones to adjust skeletal detail, while existing baselines do not provide this level of controllable output granularity.}
\label{fig:bone_count_control}
\vspace{-5mm}
\end{wrapfigure}

\textbf{Controllable Skeletal Complexity.} A unique advantage of our flow-matching framework is the ability to explicitly control skeletal density via Classifier-Free Guidance. While our model can autonomously determine a plausible bone count through unconditional generation, it also supports bone-count conditioning as a user-defined input. This allows \ourmethod to synthesize skeletons with arbitrary bone counts ranging from 1 to 128, providing direct and flexible control over the output granularity that is absent in existing autoregressive methods. As illustrated in Figure~\ref{fig:bone_count_control}, \ourmethod can adapt the skeletal detail to satisfy specific animation requirements, varying from sparse armatures for background characters to high-density rigs for complex manual dexterity, whereas other baselines remain restricted to a single, uncontrolled output density. 

\mypara{Additional Notes.} We report inference speed in Appendix~\ref{app:inference_details}, where \ourmethod is faster than all baselines. We also develop a feedforward skinning model for downstream use in Appendix~\ref{app:skin_model}, which is competitive with prior baselines.

\section{Conclusion and Future Works}
We propose \ourmethod, a unified framework for 3D and 4D skeleton rigging from static meshes and dynamic mesh sequences. Built on flow matching, \ourmethod generates temporally aligned skeletons while maintaining strong static reconstruction quality. By combining mesh-conditioned skeleton generation with a graph head for consistent topology and an uncertainty head for confidence-based sample selection, the model achieves competitive 3D performance and clear gains in 4D consistency and controllability. For future work, we plan to build stronger benchmarks for dynamic prediction under partial observations, extend the framework to jointly predict skinning weights and deformation, and introduce richer user controls such as semantic constraints or text-guided rig specification. We hope \ourmethod serves as a step toward a more general and controllable rigger for 3D and 4D content.

\bibliography{conference}
\bibliographystyle{conference}

\newpage
\appendix

\DoToC

% \newpage

%%%%%%%%%%%%%%%%%%%%%%%%%%%%%%%%%%%%%%%%%%%%%%%%%%%%%%%%%%%%%%%%%

\setcounter{figure}{0}
\setcounter{table}{0}
\setcounter{equation}{0}
\renewcommand\thesection{\Alph{section}}
\renewcommand{\thetable}{S\arabic{table}}  
\renewcommand{\thefigure}{S\arabic{figure}}
\renewcommand{\theequation}{S\arabic{equation}}

\section{Curated Dataset}\label{app:curated_dataset}

\begin{figure}[!h]
\centering
\includegraphics[width=\linewidth]{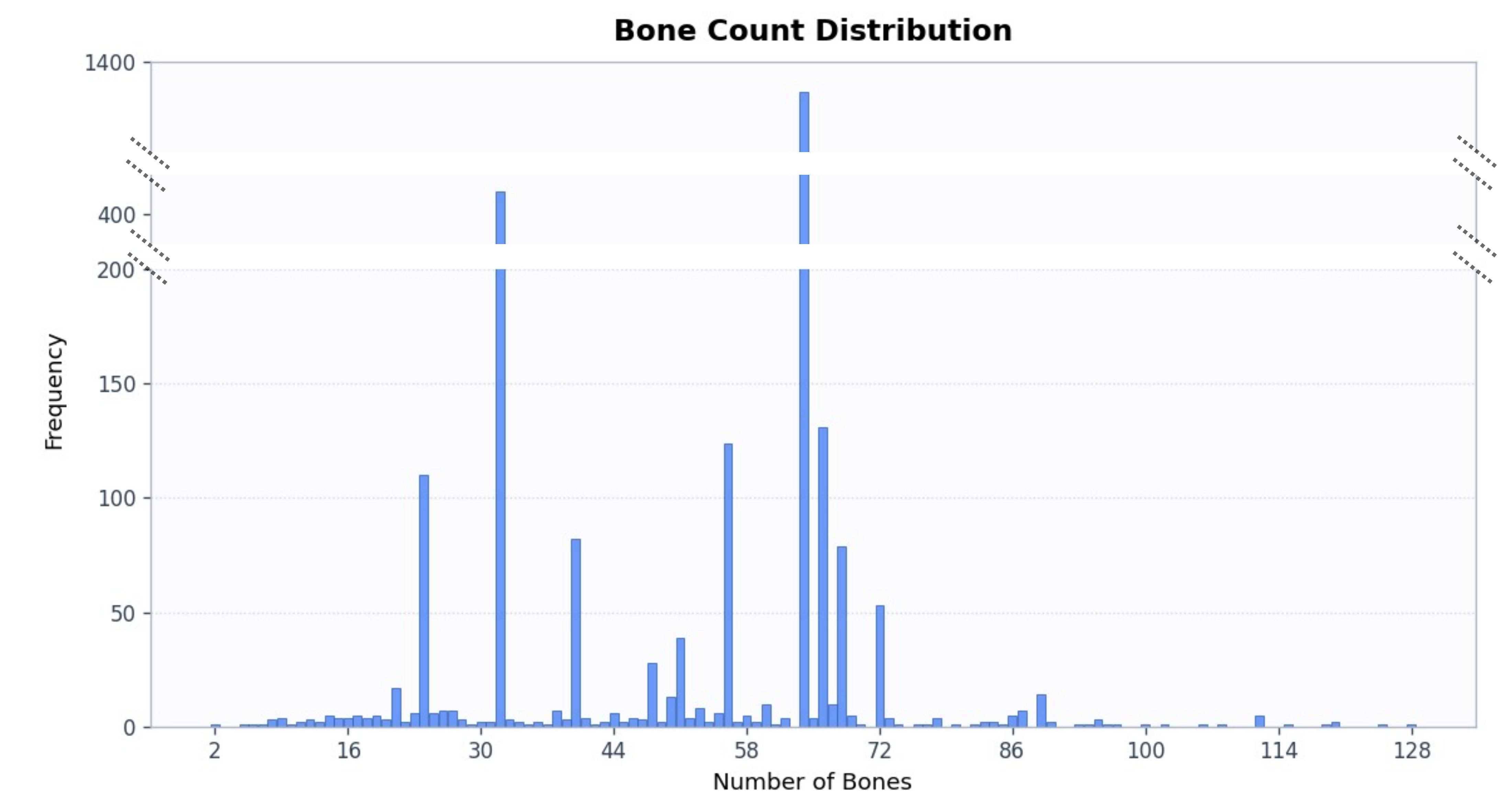}
\caption{Bone Count Histogram of the filtered TexVerse dataset.}
\label{fig:dataset}
\end{figure}

TexVerse provides a large-scale collection of animated mesh-skeleton pairs~\citep{zhang2025texverse}. However, the raw dataset is not limited to clean object-centric motions, and many sequences exhibit quasi-static motion, inconsistent temporal alignment, or incorrect skeleton annotations. To construct a reliable evaluation set, we first use Qwen3.5~\citep{qwen3.5} for preliminary filtering and then perform additional manual verification, resulting in 2,500 clean object-centric animated instances with correct mesh-skeleton alignment. The bone count distribution is shown in Figure~\ref{fig:dataset}. For training and evaluation, we divide into a train set with 2000 instances and a test set with 500 instances.

\section{Training and model details}\label{app:train_detail}

\mypara{Mesh-to-token conditioning}
We first sample $N_{p}$ surface points and normals from each input mesh to form point-cloud frames $P_i\in\mathbb{R}^{N_{p}\times 6}$. We encode each frame independently using a off-the-shelf Michelangelo point-cloud VAE to obtain a compact set of mesh tokens \citep{zhao2023michelangelo}. Concretely, we flatten the sequence to $P_{flat}\in\mathbb{R}^{N\times N_{p}\times 6}$ and pass $\{P_i\}$ independently through the point cloud encoder. The resulting conditioning tokens are reshaped to $\mathbf{C}\in\mathbb{R}^{N\times L_m\times H}$. For partially observed sequences, we mask the tokens of unobserved frames and replace with a learnable token for all masked frames.

\mypara{Model architecture.}
We use a DiT-L backbone with depth 24, hidden size $H=1152$, and 16 attention heads. Following the notation in the main method section, for each frame $i$, the input token sequence concatenates (a) $L_s$ skeleton-slot tokens obtained by linearly projecting the 7D state $\mathbf{x}_{t,i,:,:}\in\mathbb{R}^{L_s\times 7}$ to $\mathbb{R}^{L_s\times H}$ and (b) $L_m$ mesh conditioning tokens $\mathbf{C}_i\in\mathbb{R}^{L_m\times H}$ produced by the Michelangelo point-cloud VAE encoder. Here $L_s$ is the maximum number of skeleton slots, $L_m$ is the number of mesh tokens per frame, and $N$ is the sequence length. This yields a per-frame token set of size $(L_s+L_m)$ and a packed transformer stream of length $N\cdot(L_s+L_m)$. We employ an alternating attention schedule, denoted \textrm{frame-global alternating}, that interleaves frame-wise attention on frame-indexed sequences with global attention on the packed sequence, and we apply temporal RoPE using the frame index across the concatenated token stream.

\mypara{Training auxiliary loss.}
The auxiliary loss includes geometry loss and temporal smooth loss. First, we apply the geometry loss to regularize the joints outside the mesh by including a signed distance (SD) penality term:
\begin{equation}
\mathcal{L}_\textrm{geo}=\frac{1}{NL_s}\sum_{i,j} \mathrm{ReLU}(\mathrm{SD}(\hat{\mathbf{x}}_{0,i,j}, M_i))\cdot (1-t)^{\gamma_{in}},
\end{equation}
where $\gamma_{in}$ is another hyperparameter for denoising-step-aware weight.

Second, to encourage coherent dynamics and reduce temporal jitter, we additionally apply lightweight temporal regularizers on slot sequences, including bone-length consistency across adjacent frames and velocity smoothness across frame triplets:
\begin{equation}
\mathcal{L}_\textrm{smooth}=\lambda_\textrm{bl}\,\frac{1}{N-1}\sum_{i=2}^{N}\ell_\textrm{bl}(\mathbf{s}_i,\mathbf{s}_{i-1})
+\lambda_\textrm{vel}\,\frac{1}{N-2}\sum_{i=3}^{N}\ell_\textrm{vel}(\mathbf{s}_i,\mathbf{s}_{i-1},\mathbf{s}_{i-2}),    
\end{equation}

where $\mathbf{s}_i\in\mathbb{R}^{L_s\times 7}$ denotes the slot tensor at frame $i$, and $\lambda_{bl}$ and $\lambda_{vel}$ are the bone length and velocity smoothness coefficients. As a result, the total auxiliary loss is:
\begin{equation}
\mathcal{L}_\textrm{aux}=\lambda_\textrm{geo}\mathcal{L}_\textrm{geo}+\lambda_\textrm{smooth}\mathcal{L}_\textrm{smooth}
\end{equation}

\mypara{Curriculum training.}
In the first stage, we set $N=1$ and optimize the full objective with one-to-one matching. In the second stage, we train on variable-length windows and apply full-frame conditioning. In the third stage, we apply masked-conditioning mechanism:  for each training sample we draw an observed index set $\mathcal{O}$ and replace mesh tokens for unobserved frames with a mask token while injecting an observed-frame pooled summary into those positions. In practice, we increase the frequency of structured masks such as biasing the masking distribution toward structured patterns corresponding to forecasting and backcasting, including forecast masks with $\mathcal{O}=\{1,\ldots,\lfloor N/2\rfloor\}$, backcast masks with $\mathcal{O}=\{\lfloor N/2\rfloor+1,\ldots,N\}$, both-ends masks with $\mathcal{O}=\{1,N\}$, and full observation with $\mathcal{O}=\{1,\ldots,N\}$. These strategies align the training distribution with inference-time use cases without changing the underlying objective. 

\mypara{Optimization and guidance.}
We optimize the model with AdamW under mixed precision using bf16, together with cosine learning-rate decay and warmup. We set the starting learning rate to $1e-4$, $5e-5$, $3e-5$ for the three training stages of static reconstruction, dynamic reconstruction, and dynamic prediction. Skeleton-count conditioning is trained with classifier-free guidance by dropping the count label $y$ with probability $p_\textrm{drop}=0.5$, which forces the model to predict the reasonable skeleton count when no bone count guidance is provided. At inference time we apply CFG with scale $s=5$ to control the trade-off between fidelity and diversity.

\section{Evaluation Details}\label{app:eval_detail}
\subsection{Evaluation Metrics}\label{app:evaluation_metrics}

\mypara{Temporal topological consistency.}
We evaluate temporal topology with graph structure distance (GSD). Following the notation in the main method, each sequence contains $N$ frames and each frame is represented by at most $L_s$ skeleton slots. For methods without persistent slots, we first recover per-frame predicted skeletons $\{\hat{\mathbf{x}}_{0,i}\}_{i=1}^{N}$ and use the sequence-level Hungarian assignment defined in Section~3 to establish bone correspondence across time. Let $\mathcal{J}\subseteq\{1,\ldots,L_s\}$ denote the set of matched and existing bones in the anchor frame $i=1$, and let $J=|\mathcal{J}|$.

For each frame $i$, let $\pi_i(j)$ denote the parent index of matched bone $j\in\mathcal{J}$. We compare the predicted topology at frame $i$ against the anchor-frame topology $\pi_1(j)$ using the Hamming distance
\begin{equation}
H_i = \sum_{j\in\mathcal{J}} \mathbbm{1}\big[\pi_i(j) \neq \pi_1(j)\big],
\end{equation}
and normalize it to obtain the per-frame graph structure distance
\begin{equation}
\mathrm{GSD}_i = \frac{H_i}{J}.
\end{equation}
We then report the sequence-level averages
\begin{equation}
\overline{H} = \frac{1}{N-1}\sum_{i=2}^{N} H_i,
\qquad
\overline{\mathrm{GSD}} = \frac{1}{N-1}\sum_{i=2}^{N} \mathrm{GSD}_i.
\end{equation}
For fixed-slot methods, the same definition applies directly because slot correspondence is given by the slot index.

\mypara{Bone displacement deviation (BDD).}
We evaluate temporal geometric stability with bone displacement deviation (BDD). Using the same correspondence set $\mathcal{J}$, let $d_j^{(i)}$ denote the length of matched bone $j\in\mathcal{J}$ at frame $i$. We measure how much each corresponding bone deviates from its anchor-frame length and average over all matched bones and non-anchor frames:
\begin{equation}
\mathrm{BDD} = \frac{1}{J(N-1)} \sum_{j\in\mathcal{J}} \sum_{i=2}^{N} \big| d_j^{(i)} - d_j^{(1)} \big|.
\end{equation}
This correspondence-based definition ensures that BDD is computed on aligned bones rather than unmatched slots.

\subsection{Inference Details.} \label{app:inference_details}
To accurately evaluate the model performance, we sample 5 times for each instance and report the average evaluation metrics to reduce the variance of the metrics.

\begin{table}[t]
\centering
\caption{Comparison of inference time of skeleton generation on static mesh.}
\label{tab:inference_time_3d}
\begin{tabular}{lcccccc}
\toprule
Method & Pinocchio & RigNet & UniRig & MagicArti. & Puppeteer & Ours \\
\midrule
Inference time (s) & 3.9 & 4.5 & 2.9 & {2.4} & \underline{1.5} & \textbf{1.1}\\
\bottomrule
\end{tabular}
\end{table}
\begin{table}[t]
\centering
\caption{Inference time performance across different frame counts.}
\label{tab:inference_time_4D}
\begin{tabular}{l|cccccc}
\toprule
Time window size & 1 & 2 & 4 & 8 & 16 & 24 \\
\midrule
Inference Time (s) & 1.1 & 1.7 & 2.2 & 4.4 & 11.9 & 23.4 \\
\bottomrule
\end{tabular}
\end{table}

\mypara{Inference Time.} We compare average inference times among baselines. \ourmethod requires $1.1$s for single-mesh inference. As shown in Table~\ref{tab:inference_time_3d}, it is $3.5\times$ faster than Pinocchio, $4.1\times$ faster than RigNet, $2.6\times$ faster than UniRig, $2.2\times$ faster than MagicArticulate, and $1.4\times$ faster than Puppeteer. Table~\ref{tab:inference_time_4D} further shows how inference time scales with the temporal window size. These results indicate that \ourmethod remains efficient for single-frame 3D rigging while scaling reasonably to longer 4D windows.

\section{More Quantitative Rigging Results}\label{app:more_quant_results}
\begin{table}[t]
\centering
\caption{Category composition of the new Objaverse-derived 4D evaluation set. No instance of this set is used for training; the model is trained only on the TexVerse split described in \ref{app:curated_dataset}}
\label{tab:objaverse_dist}
\begin{tabular}{lcc}
\toprule
Category & $n$ & Share  \\
\midrule
bipedal mammal & 453 & 27.4\% \\
fish & 301 & 18.2\% \\
bird & 204 & 12.3\% \\
dinosaur & 193 & 11.7\% \\
quadruped mammal & 183 & 11.1\% \\
insect & 154 & 9.3\% \\
crustacean & 62 & 3.7\% \\
arachnid & 58 & 3.5\% \\
amphibian & 35 & 2.1\% \\
mollusk & 13 & 0.8\% \\
\midrule
\textbf{Total} & \textbf{1,656} & \textbf{100\%} \\
\bottomrule
\end{tabular}
\vspace{-1mm}
\end{table}
\begin{table}[t]
\centering
\small
\caption{\textbf{Per-category J2J on the Objaverse~\citep{deitke2023objaverse} 4D evaluation set.} Stratified subsample, $T = 8$. Bold is best in the row, underlined second best. MACRO averages the ten categories equally, MICRO weights by instance count.}
\label{tab:more_dy_j2j}
\begin{tabular}{lccccc}
\toprule
Category & $n$ & Puppeteer-D* & RigGS & Ours & Ours* \\
\midrule
amphibian & 20 & 0.0524 & 0.0675 & \underline{0.0389} & \textbf{0.0361} \\
arachnid & 20 & 0.0503 & 0.0567 & \underline{0.0474} & \textbf{0.0468} \\
bipedal mammal & 20 & 0.0291 & 0.0536 & \underline{0.0238} & \textbf{0.0222} \\
bird & 20 & 0.0759 & 0.0699 & \textbf{0.0627} & \underline{0.0646} \\
quadruped mammal & 20 & 0.0722 & 0.0686 & \underline{0.0564} & \textbf{0.0540} \\
crustacean & 20 & 0.0695 & 0.0637 & \underline{0.0560} & \textbf{0.0558} \\
dinosaur & 20 & 0.0644 & 0.0653 & \textbf{0.0554} & \underline{0.0619} \\
fish & 20 & 0.0733 & \textbf{0.0513} & \underline{0.0515} & 0.0525 \\
insect & 20 & 0.0692 & 0.0632 & \underline{0.0591} & \textbf{0.0571} \\
mollusk & 13 & 0.0784 & 0.0758 & \underline{0.0663} & \textbf{0.0623} \\
\midrule
\textbf{MACRO (mean of 10)} & 193 & 0.0635 & 0.0636 & \underline{0.0517} & \textbf{0.0513} \\
\textbf{MICRO (instance-weighted)} & 193 & 0.0629 & 0.0631 & \underline{0.0512} & \textbf{0.0509} \\
\bottomrule
\end{tabular}
\vspace{3mm}
% \end{table}

% \begin{table}[t]
\centering
\caption{\textbf{Per-category J2B}. Same protocol and conventions as Table \ref{tab:more_dy_j2j}.}
\label{tab:more_dy_j2b}
\begin{tabular}{lccccc}
\toprule
Category & $n$ & Puppeteer-D* & RigGS & Ours & Ours* \\
\midrule
amphibian & 20 & 0.0439 & 0.0535 & \underline{0.0289} & \textbf{0.0272} \\
arachnid & 20 & 0.0406 & 0.0423 & \underline{0.0323} & \textbf{0.0316} \\
bipedal mammal & 20 & 0.0233 & 0.0391 & \underline{0.0183} & \textbf{0.0171} \\
bird & 20 & 0.0647 & 0.0560 & \textbf{0.0507} & \underline{0.0520} \\
quadruped mammal & 20 & 0.0594 & 0.0558 & \underline{0.0432} & \textbf{0.0400} \\
crustacean & 20 & 0.0573 & 0.0475 & \underline{0.0416} & \textbf{0.0408} \\
dinosaur & 20 & 0.0553 & 0.0525 & \textbf{0.0461} & \underline{0.0523} \\
fish & 20 & 0.0589 & \textbf{0.0379} & 0.0385 & \underline{0.0381} \\
insect & 20 & 0.0583 & 0.0476 & \underline{0.0469} & \textbf{0.0444} \\
mollusk & 13 & 0.0640 & 0.0589 & \underline{0.0535} & \textbf{0.0484} \\
\midrule
\textbf{MACRO (mean of 10)} & 193 & 0.0526 & 0.0491 & \underline{0.0400} & \textbf{0.0392} \\
\textbf{MICRO (instance-weighted)} & 193 & 0.0522 & 0.0487 & \underline{0.0395} & \textbf{0.0388} \\
\bottomrule
\end{tabular}
\vspace{3mm}
% \end{table}

% \begin{table}[t]
\centering
\caption{\textbf{Per-category B2B}. Same protocol and conventions as Table \ref{tab:more_dy_j2j}.}
\label{tab:more_dy_b2b}
\begin{tabular}{lccccc}
\toprule
Category & $n$ & Puppeteer-D* & RigGS & Ours & Ours* \\
\midrule
amphibian & 20 & 0.0399 & 0.0467 & \underline{0.0258} & \textbf{0.0237} \\
arachnid & 20 & 0.0368 & 0.0369 & \underline{0.0276} & \textbf{0.0268} \\
bipedal mammal & 20 & 0.0203 & 0.0367 & \underline{0.0150} & \textbf{0.0138} \\
bird & 20 & 0.0588 & 0.0496 & \textbf{0.0437} & \underline{0.0440} \\
quadruped mammal & 20 & 0.0506 & 0.0511 & \underline{0.0372} & \textbf{0.0348} \\
crustacean & 20 & 0.0506 & 0.0421 & \underline{0.0384} & \textbf{0.0372} \\
dinosaur & 20 & 0.0480 & 0.0455 & \textbf{0.0388} & \underline{0.0453} \\
fish & 20 & 0.0453 & \underline{0.0314} & 0.0320 & \textbf{0.0307} \\
insect & 20 & 0.0529 & 0.0424 & \underline{0.0413} & \textbf{0.0383} \\
mollusk & 13 & 0.0563 & 0.0521 & \underline{0.0473} & \textbf{0.0422} \\
\midrule
\textbf{MACRO (mean of 10)} & 193 & 0.0459 & 0.0435 & \underline{0.0347} & \textbf{0.0337} \\
\textbf{MICRO (instance-weighted)} & 193 & 0.0456 & 0.0431 & \underline{0.0343} & \textbf{0.0334} \\
\bottomrule
\end{tabular}
\vspace{-1mm}
\end{table}

\subsection{Generalization and Per-Sequence Optimization Comparison} \label{app:more_4d_quant}
\mypara{Data.} To evaluate the generalization capability of \ourmethod in dynamic reconstruction, we built a second dynamic evaluation set from Objaverse~\citep{deitke2023objaverse}, a different source from the TexVerse data the model trains on, using the same VLM pre-filter, manual verification, and quality predicates detailed in Appendix \ref{app:curated_dataset}. This new dataset contains 1,656 verified animated instances across ten categories. Details of the per-category breakdown can be seen in Table \ref{tab:objaverse_dist}. 

\mypara{Per-Sequence Optimization Baseline.} To perform a more comprehensive evaluation of \ourmethod in dynamic reconstruction, we include a per-sequence optimization 4D baseline, RigGS~\citep{yao2025riggs}. This model takes video as input and produces a skeleton together with 3D Gaussians, so we render each mesh sequence to video and supply the ground-truth masks its pipeline requires. Additionally, its output lives in its own coordinate frame, so we align the generated skeleton to the ground truth in scale, position, and orientation before computing metrics. Both steps are concessions in RigGS’s favour.

\mypara{Experiment Setup.} RigGS optimizes every sequence from scratch and takes 17 minutes per 8-frame instance on an H200, so running it on the full 1,656-instance Objaverse set of W3 would cost 469 H200 GPU hours, against about 2 GPU hours for \ourmethod. We therefore sample up to 20 instances per category, giving 193 instances across the ten categories and roughly 57 H200 GPU hours. All methods are evaluated on the same subsample at $T = 8$.

\mypara{Results.}
We conclude the following based on the results in Tables \ref{tab:more_dy_j2j}, \ref{tab:more_dy_j2b}, and \ref{tab:more_dy_b2b}:
\begin{itemize}
    \item \textbf{\ourmethod leads in 28 of the 30 category-metric cells}. In aggregate, Ours* is 19.3\% below RigGS and 19.2\% below Puppeteer-D* on MACRO J2J, 20.2\% and 25.5\% below on J2B, and 22.5\% and 26.6\% below on B2B; MICRO gives the same picture. The two exceptions are fish on J2J and J2B, where RigGS leads by 0.4\% and 0.5\%, plausibly because fish motion is smooth and low in degrees of freedom, which suits per-sequence fitting.
    \item \textbf{Per-sequence optimization is not uniformly stronger than a fine-tuned static rigger.} RigGS and Puppeteer-D* are level on MACRO J2J (0.0636 vs. 0.0635), and RigGS leads only on J2B and B2B (6.7\% and 5.2\%). Optimizing each sequence individually does not by itself buy joint accuracy.
    \item \textbf{The cost gap is over two orders of magnitude.} RigGS needs 17 minutes per 8-frame instance where \ourmethod needs 4.4s in a single feedforward pass (Table \ref{tab:inference_time_4D}), roughly 230 times faster, which is what makes evaluation at the scale of the full set practical at all.
\end{itemize}

\section{More Qualitative Rigging Results}\label{app:more_results}
\begin{figure}[t]
\centering
\includegraphics[width=\linewidth]{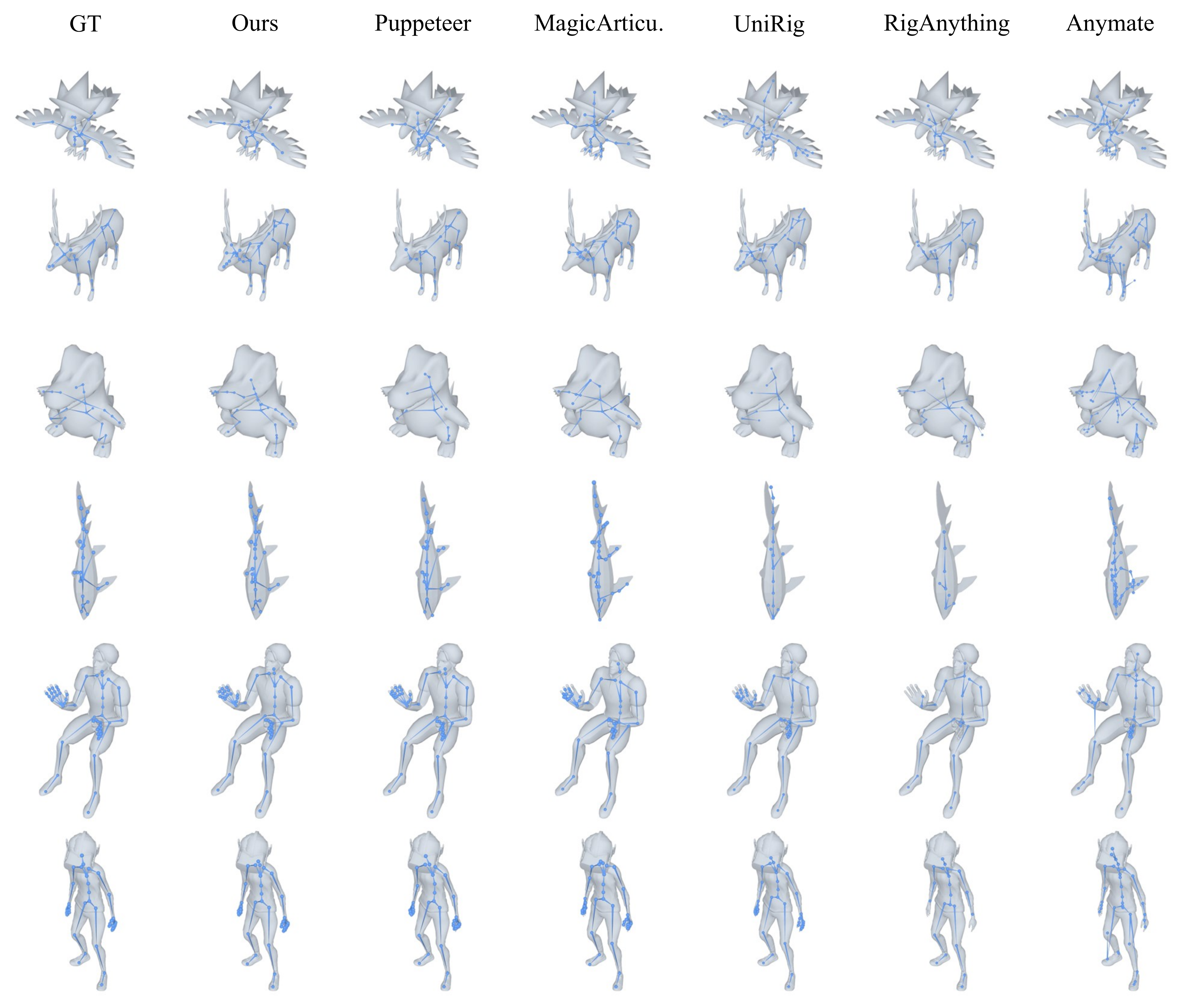}
\caption{More comparsion on static reconstruction.}
\label{fig:more_static}
\end{figure}

\begin{figure}[t]
\centering
\includegraphics[width=\linewidth]{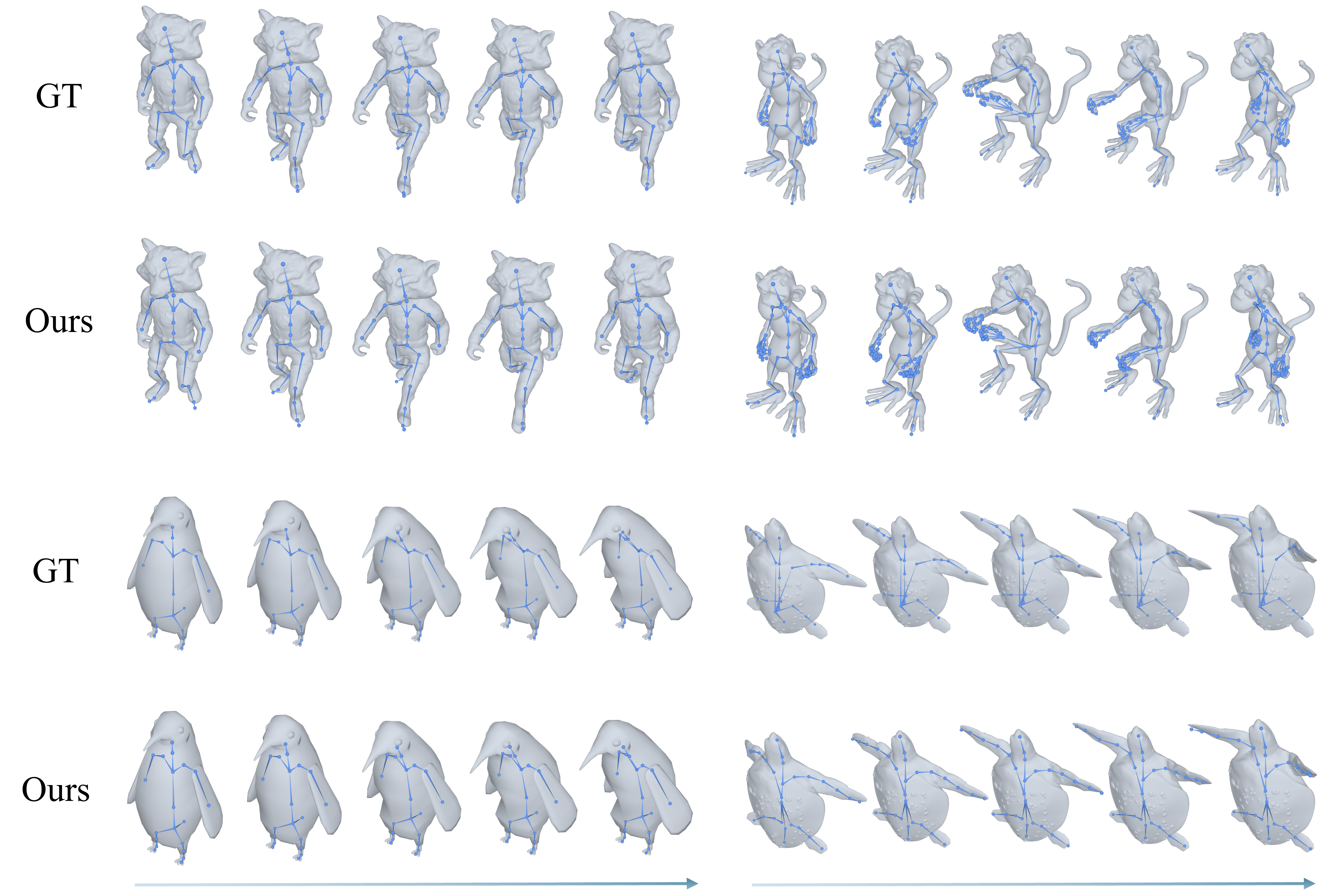}
\caption{More dynamic reconstruction results.}
\label{fig:more_dy}
\end{figure}

More static and dynamic instances are shown in Figure~\ref{fig:more_static} and Figure~\ref{fig:more_dy}, respectively.

\section{Skinning Model}\label{app:skin_model}
To animate mesh by the generated skeleton, we additionally train an attention-based skinning network, denoted \textit{SkinningTransformer}, to predict soft skinning weights from a mesh observation and its associated skeletal rig. Given a mesh and a rig with $J\leq J_{\max}$ joints, our goal is to predict, for each queried surface point $q\in\{1,\ldots,N_q\}$, a categorical distribution over valid joints. Each sample contains $N_q$ sampled query points with normals, a padded bone tensor storing parent and child joint coordinates for each bone slot, a validity mask over joint slots, and a ground-truth skinning distribution over valid joints. Formally, for a query point $q$ we model
\begin{equation}
p(\mathbf{w}_q\mid \mathcal{M},\mathbf{B},\mathbf{p}_q,\mathcal{G}),
\qquad
\mathbf{w}_q\in\mathbb{R}^{J_{\max}},
\qquad
\sum_{j=1}^{J_{\max}} w_{q,j}=1,
\end{equation}
where $\mathbf{p}_q\in\mathbb{R}^{6}$ is the queried point-and-normal feature, $\mathcal{M}$ denotes the frozen mesh encoders, and $\mathcal{G}$ denotes the skeletal graph structure.

\mypara{Frozen geometry encoders.}
We condition the skinning model on two frozen geometry pathways. First, a Michelangelo shape encoder~\citep{zhao2023michelangelo} produces a compact set of global shape tokens $\mathbf{F}_{\mathrm{shape}}\in\mathbb{R}^{257\times 768}$.
Second, the PartField encoder~\citep{liu2025partfield} produces dense query features on sampled surface locations, $\mathbf{F}_{\mathrm{pf}}^{\mathrm{surf}}\in\mathbb{R}^{N_q\times 448}$, where $N_q$ is the number of sampled surface queries and $\mathrm{PF}(\cdot)$ denotes the frozen part-aware field encoder. When skeleton-side part features are enabled, we also evaluate this field at the parent and child joint locations $\mathbf{x}^{\mathrm{par}}_j,\mathbf{x}^{\mathrm{chd}}_j\in\mathbb{R}^{3}$ of bone $j$ and average them per bone,
\begin{equation}
\mathbf{F}_{\mathrm{pf},j}^{\mathrm{skel}}=\tfrac{1}{2}\big(\mathrm{PF}(\mathbf{x}^{\mathrm{par}}_j)+\mathrm{PF}(\mathbf{x}^{\mathrm{chd}}_j)\big)\in\mathbb{R}^{448}.
\end{equation}
These encoders remain frozen throughout training.

\mypara{Shape-skeleton fusion encoder.}
All conditioning features are projected into a shared transformer width $d$. Let $\mathbf{B}\in\mathbb{R}^{J_{\max}\times 6}$ denote the padded bone tensor, where each row stores the concatenated parent and child joint coordinates of one bone. The learned projection matrices satisfy $\mathbf{W}_s\in\mathbb{R}^{768\times d}$, $\mathbf{W}_j\in\mathbb{R}^{6\times d}$, $\mathbf{W}^{sp}\in\mathbb{R}^{448\times d}$, and $\mathbf{W}_p\in\mathbb{R}^{448\times d}$. We form projected shape and bone tokens as
\begin{equation}
\mathbf{F}'_{\mathrm{shape}}=\mathbf{F}_{\mathrm{shape}}\mathbf{W}_s,
\qquad
\mathbf{H}^{(0)}_{\mathrm{bone}}=\mathbf{B}\mathbf{W}_j+\boldsymbol{\mathcal{E}}^{pos}_{:J_{\max}}+[\mathbf{F}_{\mathrm{pf}}^{\mathrm{skel}}\mathbf{W}^{sp}],
\end{equation}
where $\boldsymbol{\mathcal{E}}^{pos}_{:J_{\max}}\in\mathbb{R}^{J_{\max}\times d}$ is a learnable positional embedding for bone slots and the last term is optional. The encoder input is then
\begin{equation}
\mathbf{Z}^{(0)}=\mathrm{Concat}(\mathbf{F}'_{\mathrm{shape}},\mathbf{H}^{(0)}_{\mathrm{bone}})
\in\mathbb{R}^{L_{\mathrm{enc}}\times d},
\qquad
L_{\mathrm{enc}}=257+J_{\max}.
\end{equation}
A stack of transformer encoder blocks updates $\mathbf{Z}$, while invalid joint slots are masked so that padded joints do not participate in attention.

\mypara{Topology-aware joint attention.}
To encode skeletal structure explicitly, we inject a topology-aware bias into self-attention over the bone tokens \citep{song2025puppeteer}. Let $\mathbf{D}\in\mathbb{R}^{J_{\max}\times J_{\max}}$ denote the pairwise graph-distance matrix on the skeleton. Distances are clamped into a small set of buckets and projected into per-head biases. The attention score for head $h$, with query/key head dimension $d_h$, becomes
\begin{equation}
\mathrm{score}^{(h)}_{ij}=
\frac{\mathbf{Q}^{(h)}_i {\mathbf{K}^{(h)}_j}^{\top}}{\sqrt{d_h}}
+\lambda_{\mathrm{taja}}\, b^{(h)}_{ij}(\mathbf{D}),
\end{equation}
for bone-token indices $i$ and $j$. This topology-aware joint attention encourages joints that are nearby in the kinematic tree to exchange information more strongly.

\mypara{Point-wise cross-attention decoder.}
Surface query tokens are initialized from projected part-aware point features,
\begin{equation}
\mathbf{H}^{(0)}_{\mathrm{pt}}=\mathbf{F}^{\mathrm{surf}}_{\mathrm{pf}}\mathbf{W}_p\in\mathbb{R}^{N_q\times d}.
\end{equation}
Optionally, Fourier positional encodings of the queried coordinates and normals are added to these tokens. The decoder then refines point tokens purely through cross-attention to the encoder memory. There is no decoder self-attention; interaction among queries is mediated through the shared shape-skeleton memory.

\mypara{Scaled cosine prediction head.}
The final logits are computed by cosine similarity between normalized point embeddings and normalized joint embeddings, scaled by a learnable factor $\alpha$, where $z_{qj}$ is the logit for assigning query point $q$ to joint $j$,
\begin{equation}
z_{qj}=(|\alpha|+\varepsilon)
\cdot
\frac{\mathbf{F}^{\mathrm{out}}_{\mathrm{pt}}(q)}{\|\mathbf{F}^{\mathrm{out}}_{\mathrm{pt}}(q)\|_2}
\cdot
\frac{\mathbf{F}^{\mathrm{joint}}_{\mathrm{bone}}(j)}{\|\mathbf{F}^{\mathrm{joint}}_{\mathrm{bone}}(j)\|_2}.
\end{equation}
Invalid joint logits are masked before normalization, and skinning probabilities are obtained by
\begin{equation}
\mathbf{w}_q=\mathrm{softmax}(\mathbf{z}_q).
\end{equation}
This design makes the prediction explicitly probabilistic and naturally handles variable valid-joint sets through masking.

\mypara{Training objective.}
We optimize the skinning network with a masked soft cross-entropy objective together with an $L_1$ loss in probability space. Let $\mathcal{Y}$ denote the target weight distributions and let $\mathbf{Z}$ denote the predicted logits. The overall loss is
\begin{equation}
\mathcal{L}=\mathcal{L}_{\mathrm{CE}}(\mathcal{Y},\mathbf{Z};\mathbf{m})
+\lambda_{\ell_1}\,\mathcal{L}_{L_1}(\sigma(\mathbf{Z}),\mathcal{Y};\mathbf{m}),
\end{equation}
where $\mathbf{m}\in\{0,1\}^{N_q\times J_{\max}}$ is the joint-validity mask, $\sigma(\cdot)$ denotes the row-wise softmax operation, and $\lambda_{\ell_1}$ is the weight on the probability-space $L_1$ loss. Only the SkinningTransformer parameters are updated during training; all mesh encoders remain frozen.

\begin{table}[t]
\centering
\caption{Quantitative comparison of skinning weight prediction. Precision (Prec.) and Recall (Rec.) quantify the accuracy and coverage of joint influence regions, respectively, where higher values ($\uparrow$) denote superior performance. Numerical fidelity is assessed via the mean $L_1$ error, where lower values ($\downarrow$) indicate closer alignment with the ground truth.}
\vspace{-3mm}
\label{tab:skinning_weight_comparison}
\resizebox{\textwidth}{!}{%
\begin{tabular}{lccccccccc}
\toprule
\multirow{2}{*}{Method} & \multicolumn{3}{c}{Articulation-XL2.0} & \multicolumn{3}{c}{ModelsResource} & \multicolumn{3}{c}{Diverse-pose} \\ \cmidrule(lr){2-4} \cmidrule(lr){5-7} \cmidrule(l){8-10} 
 & Prec. $\uparrow$ & Rec. $\uparrow$ & L1 $\downarrow$ & Prec. $\uparrow$ & Rec. $\uparrow$ & L1 $\downarrow$ & Prec. $\uparrow$ & Rec. $\uparrow$ & L1 $\downarrow$ \\ \midrule
RigNet & 73.7\% & 66.1\% & 0.729 & 65.7\% & 80.2\% & 0.707 & 74.7\% & 65.4\% & 0.746 \\
MagicArti. & 74.6\% & 71.3\% & 0.451 & 68.1\% & 80.7\% & 0.642 & 74.9\% & 68.4\% & 0.479 \\
Puppeteer  & 87.6\% & \underline{74.0}\% & 0.335 & 79.7\% & \textbf{81.6\%} & \underline{0.443} & 83.6\% & 72.2\% & 0.405 \\
Puppeteer-D & \underline{87.9\%} & 73.8\% & \underline{0.333} & \underline{79.8\%} & \underline{81.5}\% & \textbf{0.442} & \underline{86.4\%} & 72.8\% & \underline{0.353} \\ 
Ours  & \textbf{89.7\%} & \textbf{75.0\%} & \textbf{0.311} & \textbf{81.8\%} & 76.2\% & 0.449 & \textbf{88.3\%} & \textbf{73.6\%} & \textbf{0.348} \\
\bottomrule
\end{tabular}%
}
\end{table}

\begin{figure}[t]
\centering
\includegraphics[width=0.6\linewidth]{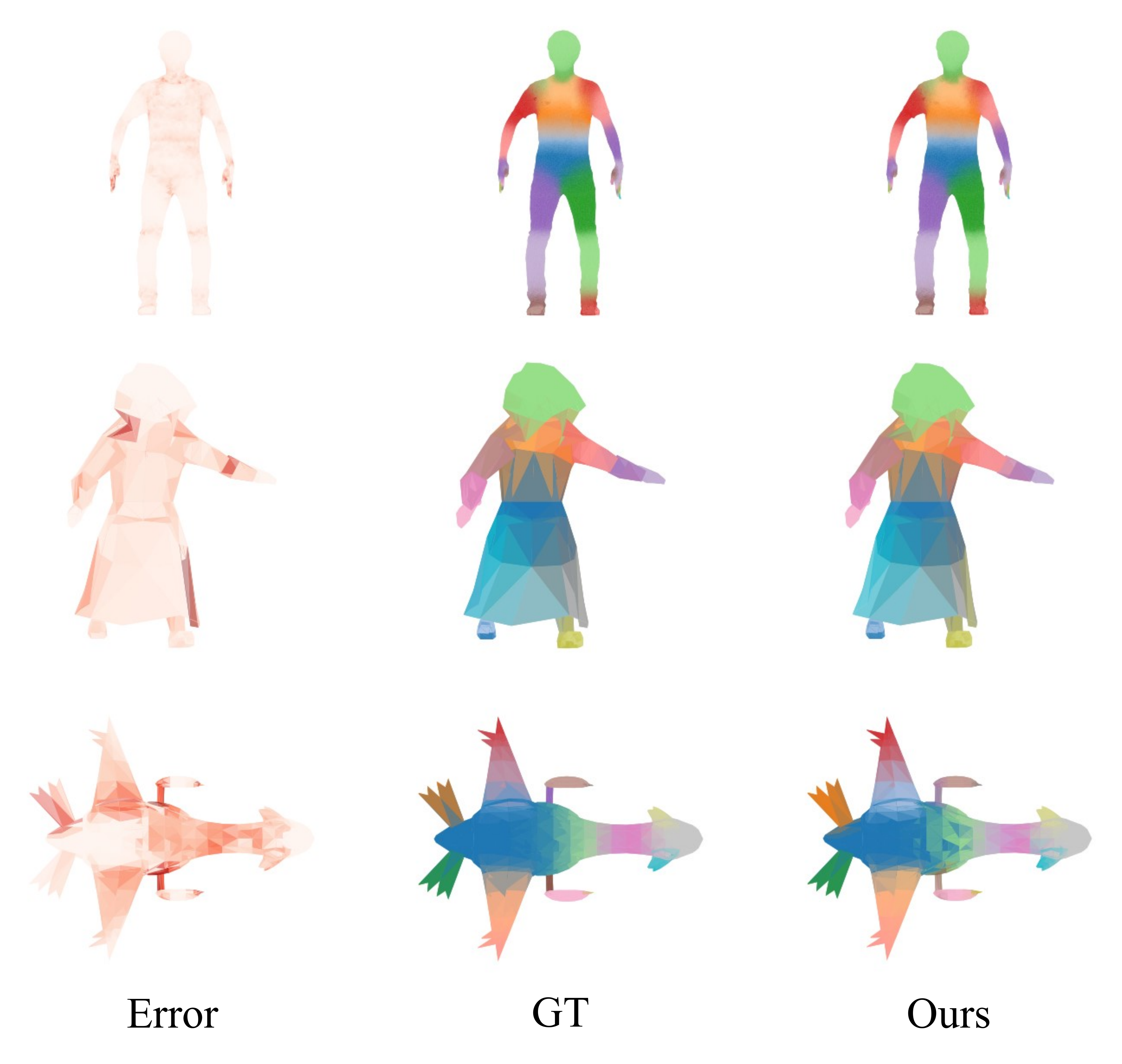}
\caption{Qualitative results of our skinning model.}
\label{fig:skin}
\end{figure}

\mypara{Evaluation protocol.}
At test time, predictions on sampled query points are transferred to mesh vertices through nearest-neighbor interpolation or Gaussian-weighted $k$-NN blending. Optional post-processing suppresses low-confidence weights and enforces a sparse set of active joints per vertex. This produces the final per-vertex skinning weights used for downstream deformation and evaluation.

\mypara{Experiments. }
Table~\ref{tab:skinning_weight_comparison} shows that our skinning model achieves the strongest overall performance across the evaluated datasets. On Articulation-XL2.0 and Diverse-pose, it obtains the best precision, recall, and L1 error, indicating more accurate and more complete skinning weight prediction. On ModelsResource, our method achieves the best precision and remains competitive in L1 error. Overall, these results suggest that the proposed skinning model generalizes well across diverse object categories and preserves strong prediction quality in both in-domain and out-of-domain settings. Figure~\ref{fig:skin} shows qualitative results.

\end{document}